\documentclass{article} 
\usepackage{iclr2026_conference,times}

\usepackage{amsmath,amsfonts,bm}

\def\eqref#1{equation~\ref{#1}}

\def\1{\bm{1}}

\DeclareMathAlphabet{\mathsfit}{\encodingdefault}{\sfdefault}{m}{sl}
\SetMathAlphabet{\mathsfit}{bold}{\encodingdefault}{\sfdefault}{bx}{n}

\DeclareMathOperator*{\argmax}{arg\,max}

\usepackage{marvosym}
\usepackage{hyperref}
\usepackage{url}
\usepackage{wrapfig}
\usepackage{times}  
\usepackage{helvet}  
\usepackage{courier}  
\usepackage{graphicx} 
\usepackage{natbib}  
\usepackage{booktabs}       
\usepackage{multirow}       
\usepackage{colortbl}       
\usepackage{tabularx}
\usepackage[table]{xcolor}  
\usepackage{graphicx}       
\definecolor{gray}{gray}{0.9}  
\usepackage{adjustbox}
\usepackage{caption} 

\newcommand{\positive}[1]{\textcolor{red}{$\uparrow$#1}}
\definecolor{mygreen}{rgb}{0.0,0.6,0.0}
\newcommand{\negative}[1]{\textcolor{mygreen}{$\downarrow$#1}}
\definecolor{toponeRed}{HTML}{FF9999} 
\definecolor{toptwoRed}{HTML}{FFCCCC} 
\usepackage{pifont}
\usepackage{float}
\newcommand{\xmark}{\ding{55}}
\usepackage{algorithm}
\usepackage{algorithmic}
\usepackage{amsfonts}
\usepackage{amsmath}
\usepackage[most]{tcolorbox}
\usepackage{newfloat}
\usepackage{listings}

\title{CTTS: Collective Test-Time Scaling}

\author{
Zhende Song$^{1,2}$ Shengji Tang$^{2,3}$ Peng Ye$^{2,\textrm{\Letter}}$ Jiayuan Fan$^{1,\textrm{\Letter}}$ Lei Bai$^{2}$ Tao Chen$^{1}$ Wanli Ouyang$^{2,3}$
\\
  $^1$Fudan University \quad $^2$Shanghai AI Laboratory \quad $^3$The Chinese University of Hong Kong
}

\iclrfinalcopy 
\begin{document}

\maketitle

\begin{abstract}

Test-time scaling (TTS) has emerged as a promising, training-free approach for enhancing large language model (LLM) performance. However, the efficacy of existing methods, such as Best-of-N and Self-Consistency, is fundamentally constrained by the dominant single test-time scaling (STTS) paradigm, which relies on a single LLM agent interacting with a single reward model (SA-SR). Inspired by recent work showing that collective methods can surpass the performance ceiling of individual models, we introduce \textbf{Collective Test-Time Scaling (CTTS)}.
First, we systematically investigate three primary interaction paradigms of existing multiple models: single-agent-multi-reward (SA-MR), multi-agent-single-reward (MA-SR), and multi-agent-multi-reward (MA-MR). Extensive experiments reveal that the MA-MR paradigm is consistently superior. Based on this finding, we further propose \textbf{CTTS-MM}, a novel framework that operationalizes multi-agent and multi-reward collaboration. CTTS-MM integrates two key technical contributions: (1) for agent collaboration, an \textbf{Agent Collaboration Search (ACS)} that identifies the most effective combination of LLMs from a candidate pool; and (2) for reward model collaboration, a \textbf{Mixture of Reward Models (MoR)} strategy that leverages a Prior Reward model Ensemble Selection (PRES) algorithm to select the optimal ensemble.
Evaluations across seven mainstream benchmarks demonstrate that CTTS-MM significantly outperforms leading STTS methods (\textbf{+4.82\% over Best-of-N}) and surpasses even flagship proprietary LLMs (\textbf{+7.06\% over GPT-4.1}) and open-source LLMs. These results highlight the substantial potential of collective scaling to push the frontier of LLM inference.
\end{abstract}

\section{Introduction}
Recent advancements in large language models (LLMs)~\cite{gpt-4.1,Yang2024Qwen25TR,Brown2020LanguageMA,DeepSeekAI2025DeepSeekR1IR,Touvron2023LLaMAOA} have marked a significant milestone in natural language understanding and generation. LLMs are typically optimized through training-time scaling, where huge amounts of data and parameters are applied, facing growing limitations due to their resource-intensive nature and the endless hunger for human data. To avoid introducing an extra expensive training process, test-time scaling (TTS) has emerged as an orthogonal direction for fully stimulating the ability of pre-trained LLMs during inference. The process of typical TTS methods~\cite{bestofn,wang2023selfconsistency,monkey,madaan2023selfrefine}, i.e., self-repetition-based methods~\cite{bestofn,monkey} can be divided into two sequential stages: 1) an LLM agent generates multiple candidate answers; 2) an external selector (reward model or manually designed selection metric) chooses the best answer. The performance of TTS highly relies on the inference quality of the LLM agent and the selection accuracy of the selector. Currently, TTS methods primarily adopt single TTS paradigm that consists of a single agent with a single selector (referred to as a ``single to single" framework), which introduces two major limitations: 1) In the first stage, the limited capacity of a single agent causes a biased output distribution, leading to a constrained performance upper bound; 2) In the second stage, it imposes a prior selection preference, which hinders comprehensive and high-quality scoring of candidate answers. These intrinsic limitations of the ``single to single" framework impede the further performance improvement of TTS and even lead to collapse. Thus, an essential question naturally arises: \textbf{How can TTS overcome the ``single to single" framework to release the potential of LLM inference?}

\begin{figure*}[!th]
  \centering
  \includegraphics[width=\textwidth]{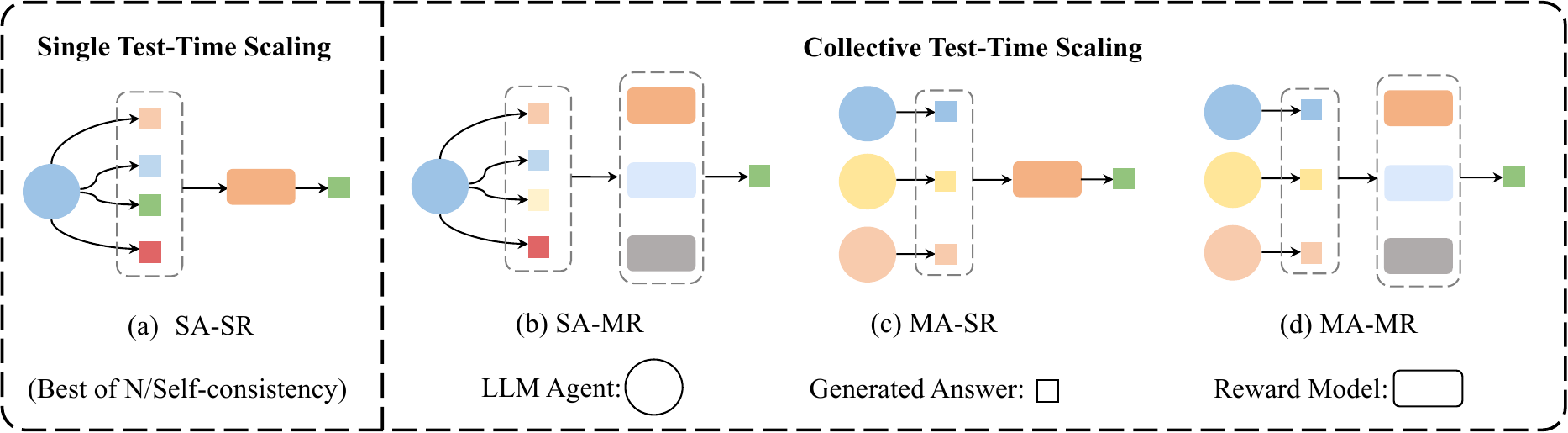}
  \caption{Comparison of previous Single Test-time Scaling (STTS) paradigm and our proposed novel Collective Test-time Scaling (CTTS) paradigms.}
  \vspace{-15pt}
  \label{fig:teaser}
\end{figure*}

Human behavior may offer some insights into the question. When tackling problems, people often engage in collaboration within teams to reach better solutions. Further, particularly challenging tasks \begin{wrapfigure}{r}{0.45\textwidth}
    \centering
    \includegraphics[width=0.45\textwidth]{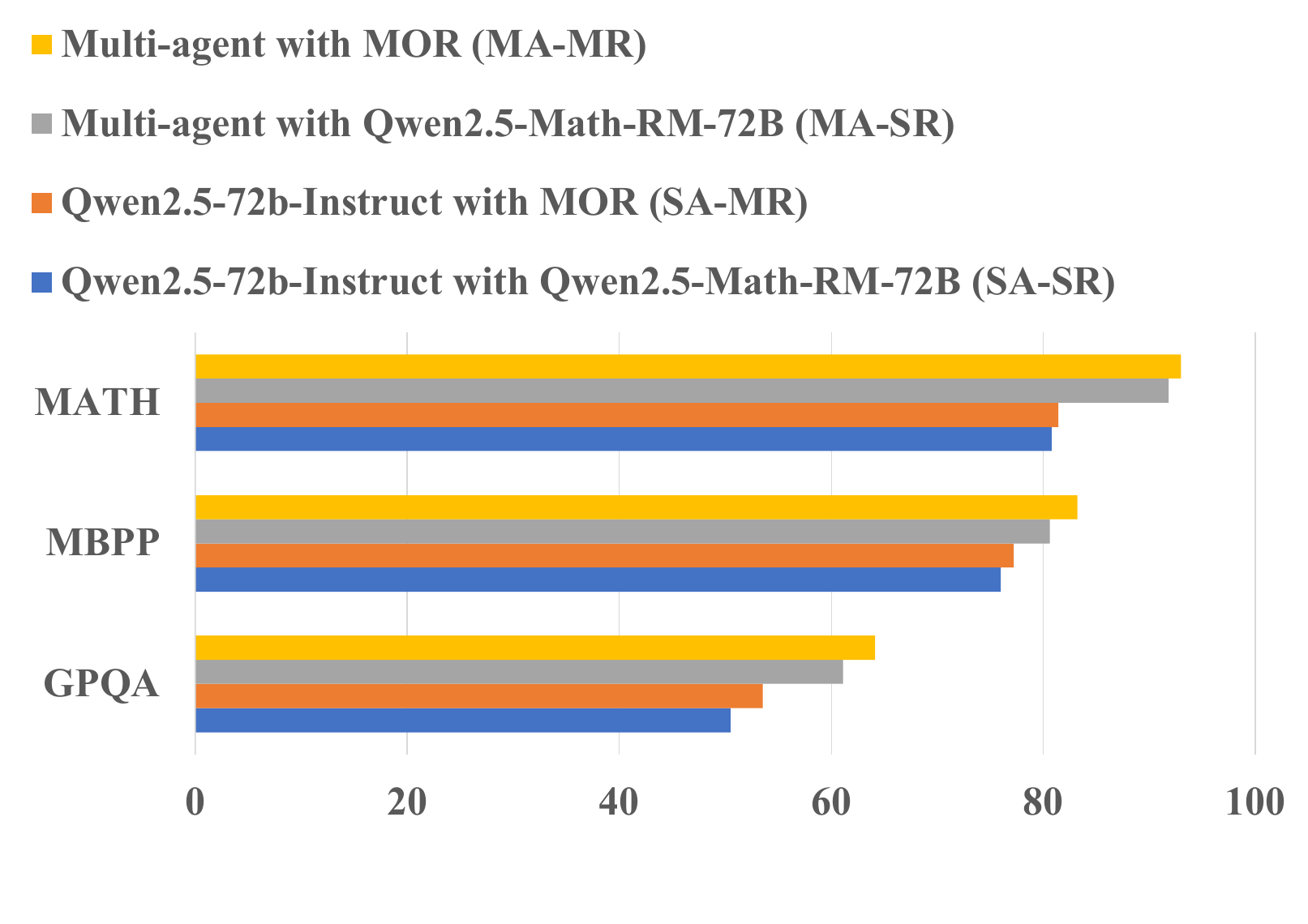}
    \caption{Comparison of three CTTS paradigm and one STTS paradigm on MATH, MBPP and GPQA.}
    \label{fig:stat}
\end{wrapfigure}may require cooperation across multiple groups, combining diverse perspectives to aggregate a more comprehensive and effective outcome. This pattern is also reflected in recent developments of collective methods~\cite{wang2025mixtureofagents,Chen2025SymbolicMA,shnitzer2024large,lu-etal-2024-routing}. For instance, Mixture-of-Agents (MoA)~\cite{wang2025mixtureofagents} exploits the references from diverse LLM agents to aggregate a higher-quality final answer, breaking through the upper bound of single-agent systems. Inspired by collective methods, we advance the previous \textbf{Single TTS (STTS)} to the novel \textbf{Collective Test-Time Scaling (CTTS)}, taking the first step to explore the potential of CTTS. Specifically, we focus on investigating two key questions:
 (1) What is the optimal paradigm of collective test-time scaling?
 (2) How can we effectively scale the systems under such paradigm?
To address the two questions above, we systematically design and explore three CTTS paradigms:
 (1) \textbf{single agent to multiple reward models (SA-MR)};
 (2) \textbf{multiple agents to single reward model (MA-SR)};
 (3) \textbf{multiple agents to multiple reward models (MA-MR)}.
Figure~\ref{fig:teaser} illustrates the differences between our proposed CTTS paradigms and the existing single TTS paradigm. To obtain the optimal CTTS paradigm, we conduct experiments to compare the above four paradigms under three benchmarks. The results are shown in Figure~\ref{fig:stat}. It can be observed that, as the collective level increases, the performance improves, and the MA-MR paradigm consistently achieves the most substantial performance improvements, highlighting that both multi-agent and multi-reward-model collaboration play a critical role in the LLM inference performance.

Based on this observation, we adopt CTTS with MA-MR paradigm as the basic framework and propose a novel CTTS method called \underline{C}ollective \underline{T}est-\underline{T}ime \underline{S}caling with \underline{M}ultiple agents to \underline{M}ultiple reward models (\textbf{CTTS-MM}) as an effective and simple specific instance of MA-MR paradigm. Specifically, for multi-agent collaboration specific to TTS, we first employ an \textbf{Agent Collaboration Search (ACS)} to choose the most effective agent ensemble from a candidate model pool. To guide the search with high-quality feedback, we propose a \textbf{Mixture of Reward Models (MoR)} to achieve multi-reward-model collaboration that breaks through the upper bound of a single reward model. To construct MoR regarding the given question, Prior Reward model Ensemble Selection (PRES) is proposed to select the optimal reward model or a weighted combination of them based on Pair-wise Reward Ranking (PRR) metric over a curated question pool. 
To verify the effectiveness of our proposed CTTS-MM, we conduct extensive experiments on seven mainstream benchmarks with ten open-source LLM agents and eight reward models. Compared with existing popular TTS, collaboration methods and leading LLMs, CTTS-MM achieves significant superiority. For instance, CTTS-MM remarkably outperforms Self-consistency by 7.68\% and Best of N by 4.83\%. Moreover, by only utilizing open-source models, CTTS surpasses flagship closed-source LLMs, including GPT-4.1 and Claude-3.7-sonnet, which demonstrates CTTS-MM can fully release the potential of models during inference time. Our contribution can be summarized as follows:

\begin{itemize}
  \item We take the first step towards formalizing and analyzing different paradigms of Collective Test-Time Scaling (CTTS), including 1) single agent to multiple reward models (SA-MR); 2) Multiple agents to Single reward model (MA-SR); and 3) multiple agents to multiple reward models(MA-MR). Our study reveals that MA-MR is the optimal CTTS paradigm due to the intra- and inter-collaboration of model groups.

  \item 
  
  We propose a novel CTTS framework named CTTS-MM, which combines multiple LLM agents and multiple reward models in a unified search–reward–search pipeline. Specifically, Agent Collaboration Search (ACS) is proposed to dynamically select an optimal combination of agents from a candidate pool, and Mixture of Reward models (MoR) is proposed to provide high-quality feedback. To achieve MoR, a Prior Reward model Ensemble Selection (PRES) with a Pair-wise Reward Ranking (PRR) metric is designed to construct an effective and adaptive reward signal.

  \item Extensive experiments across multiple benchmarks demonstrate that our CTTS-MM
  consistently outperforms existing STTS methods (+6.02\% over Symbolic-MoE, +7.09\% over MoA),  leading proprietary LLMs (+7.06\% over GPT-4.1) and various open-source LLMs, 
  
  validating the effectiveness of the proposed CTTS-MM framework and highlighting the substantial potential of collective test time scaling.
  
\end{itemize}

\section{Related Work}

\paragraph{Test-Time Scaling}
Test-time scaling methods~\cite{bestofn,monkey,madaan2023selfrefine,wang2023selfconsistency,debate,wei2022chain,yao2023tree,chen2024universal} mainly focus on how to enhance LLM agents' capabilities at test time. Best of N~\cite{bestofn} is a classic TTS approach that generate answers multiple times with LLM agents and obtains the best answer based on the reward score. Similar methods~\cite{chen2024universal} called self-consistency essentially follow the same paradigm, except that they use a verifier to select the answer. This verifier can be an evaluation tool or an algorithm like majority voting. Self-refine~\cite{madaan2023selfrefine} obtains the optimal solution through a self-evaluation and self-correction approach while ~\cite{debate} employs a multi-round debating between two agents to reach the final answer.

\paragraph{Multi-agent Collaboration}
A growing number of researches have explored collaborative strategies among multiple agents. Emerging research~\cite{Chen2025SymbolicMA,lu-etal-2024-routing,shnitzer2024large,srivatsa-etal-2024-harnessing,wang2025mixtureofagents} aims to make selection decisions before response generation, directing queries to appropriate agents in advance. MoA~\cite{wang2025mixtureofagents} exemplifies this by assigning LLM agents into an ensemble system. Symbolic-MoE~\cite{Chen2025SymbolicMA} proposes a Mixture-of-Experts framework that dynamically selects and
combines agents based on skill-specific expertise. Other methods~\cite{chen2024are,li2024more,chen2024frugalgpt,gui2024bonbon,wang2023selfconsistency} fuse the results of multiple model outputs to yield a refined answer.

\section{Methodology}

\begin{figure*}[t]
  \centering
  \includegraphics[width=\textwidth]{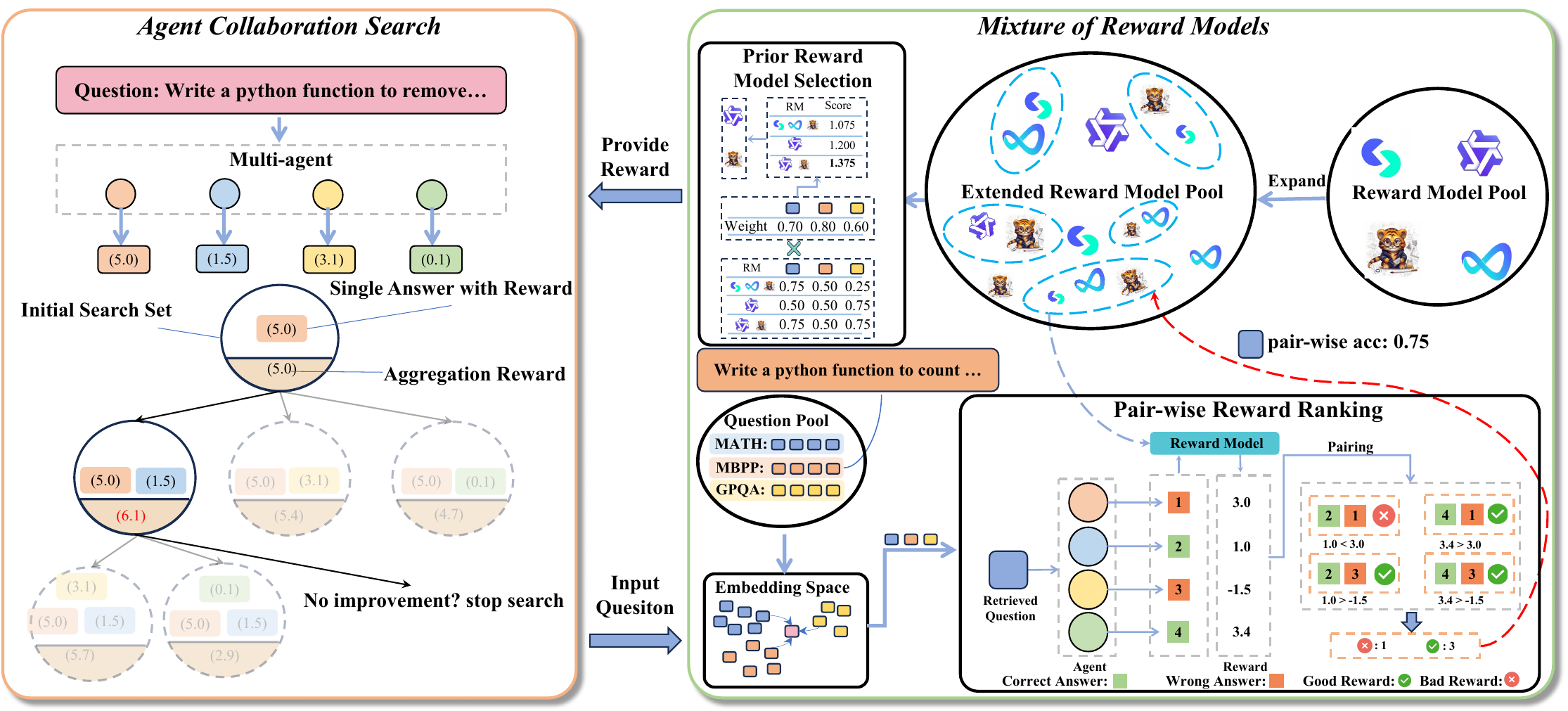}
  \caption{Overview of the proposed CTTS-MM framework. The left part illustrates the Agent Collaboration Search (ACS) while the right part depicts the Mixture of Reward Models (MoR).}
  \vspace{-8pt}
  \label{fig:methode}
\end{figure*}

In this section, we first provide a brief preliminary to elaborate on the specific framework of three CTTS paradigms. Then we introduce our proposed CTTS-MM. In Section~\ref{sec:ACS}, we introduce our Agent Collaboration Search (ACS). Section~\ref{sec:mor} details our proposed Mixture of Reward model (MoR) for selecting the optimal combination of reward models. The construction of a question pool for later selection is first presented. We then introduce Pair-wise Reward Ranking (PRR) and Prior Reward Model Ensemble Selection (PRES). Overall framework is illustrated in Figure~\ref{fig:methode}

\subsection{Preliminary}
Figure~\ref{fig:teaser} illustrates three CTTS paradigms we aim to explore: (1) single agent to multiple reward models (SA-MR); (2) multiple agents to single reward model (MA-SR); (3) multiple agents to multiple reward models (MA-MR). We design a search-reward framework to systematically investigate all three paradigms. For the specific framework setting of each paradigm, MA-SR performs multi-agent ACS with a single reward model while SA-MR adopts ACS using a single agent with MoR. Note that for SA-MR, ACS is performed under multiple answers generated by a single agent. Finally, MA-MR builds upon the previous two paradigms by jointly leveraging ACS and MoR.

\subsection{Agent Collaboration Search}
\label{sec:ACS}

The process of ACS is illustrated in Figure~\ref{fig:methode}. We design ACS based on a simple yet effective greedy search algorithm with early stop and residual aggregation. Specifically, given a question $q$, we first collect $n$ candidate answers from $n$ agents (under SA-MR setting, $n$ candidates come from repeated generation of one agent), denoted as $\mathcal{A} = \{A_0, A_1, \dots, A_{n-1}\}$. Our goal is to obtain the optimal answer from these candidates through an iterative and reward-guided greedy search. We begin by computing the reward score for each candidate using Mixture of Reward Models, denoted as function $\mathrm{MoR}$. Specifically, for each answer $A_i$, we obtain its reward score by:
\begin{equation}
    r_i = \mathrm{MoR}(q, A_i), \quad i = 0, 1, \dots, n-1.
\end{equation}
We then sort the candidates based on their scores and select the top-$k$ answers to initialize our search set $\mathcal{S}^{(0)}$:
\begin{equation}
    \mathcal{S}^{(0)} = \{A_{(0)}, A_{(1)}, \dots, A_{(k-1)}\},
\end{equation}
where $A_{(i)}$ denotes the $i$-th ranked answer by score. An aggregator $Agg$ is then used to summarize the current set of answers into a single composite response:
\begin{equation}
    C^{opt} = \mathrm{Agg}(\mathcal{S}^{(0)}),
\end{equation}
and its corresponding reward score is computed as:
\begin{equation}
    r^{opt} = \mathrm{MoR}(q, C^{opt}).
\end{equation}
where $C^{opt}$ and $r^{opt}$ are the current optimal answer and its corresponding reward score, respectively. 

We then iteratively check whether augmenting the initial search set $\mathcal{S}^{(0)}$ with a remaining candidate $A_j \in \mathcal{A} \setminus \mathcal{S}^{(0)}$ can yield a better answer. For each such candidate $A_j$, we compute:
\begin{equation}
    \hat{C}_j = \mathrm{Agg}(\mathcal{S}^{(0)} \cup \{A_j\}),
\end{equation}
\begin{equation}
    \hat{r}_j = \mathrm{MoR}(q, \hat{A}_j).
\end{equation}
We identify the candidate $A_j^*$ that yields the highest reward score:
\begin{equation}
    A_j^* = \argmax_{A_j \in \mathcal{A} \setminus \mathcal{S}^{(0)}} \hat{r}_j.
\end{equation}
If $\hat{r}_{j^*} > r^{opt}$, we update:
\begin{equation}
\left\{
\begin{aligned}
    \mathcal{S}^{(1)} &= \mathcal{S}^{(0)} \cup \{A_j^*\}, \\
    C^{opt} &= \hat{C}_{j^*}, \\
    r^{opt} &= \hat{r}_{j^*}
\end{aligned}
\right.
\end{equation}
and repeat the process using $\mathcal{S}^{(1)}$ as the new base set. Otherwise, if no such improvement is found, the search terminates and $C^{opt}$ is taken as the current optimal answer. Moreover, to mitigate potential information loss during greedy search, we incorporate a residual aggregation step. Specifically, we aggregate the final optimal answer $C^{opt}$ with the initial candidate set $\mathcal{A}$ to produce:
\begin{equation}
    \left\{
    \begin{aligned}
    C^{\text{res}} &= \mathrm{Agg}(\mathcal{A} \cup \{C^{opt}\}), \\
    r^{\text{res}} &= \mathrm{MoR}(q, C^{\text{res}})
    \end{aligned}
    \right.
\end{equation}
If $r^{\text{res}} > r^{opt}$, we replace $C^{opt}$ with $C^{\text{res}}$ as the final output. Otherwise, we keep the original output.

\subsection{Mixture of Reward Model}
\label{sec:mor}

The multi-reward-model system aims to provide accurate reward scores for the preceding greedy search process. The key challenge lies in selecting suitable reward models for different questions since reward models are currently very domain-specific. Existing approaches~\cite{bestofn} typically rely on manually selecting specific reward models tailored to specific datasets or domains. While such methods may perform well on particular datasets, they lack generalization and flexibility. We argue that this challenge is fundamentally aligned with the motivation behind MoA, which aims to enhance performance and generalization through complementary collaboration among diverse agents. The essence of MoR is somewhat analogous, which is enhancing the precision of the provided rewards through the interaction and collaboration of reward models across different domains. Our core idea is to first expand the individual reward models by constructing a reward model pool and combining them through different subsets of reward models using various weighting methods. This approach allows us to significantly extend the capacity of the original pool. The next step is to select the most suitable individual reward models or their weighted combinations from the pool. Inspired by Retrieval-Augmented Generation (RAG) methods~\cite{retrieval,benchret}, we introduce a diversified question pool as a prior for selecting the best reward model(s). We then propose a novel ranking metric called Pair-wise Reward Ranking (PRR) to evaluate the capability of reward models in assessing outputs from LLM agents. Based on the above techniques, a prior based reward model selection method named Prior Reward Model Ensemble Selection (PRES) is proposed. In this section, we first describe how the question pool is constructed, followed by the introduction of PRR. Finally, the details of PRES are presented.

\subsubsection{Diversified Question Pool}
As mentioned, manual selection of reward model(s) based on the domain of the dataset is neither generalizable nor flexible. On the other hand, it is difficult to directly select reward model(s) based on their architectures or parameters. To address this, we introduce a diversified question pool as a form of prior knowledge to guide the selection process. We construct the question pool using the validation sets of diverse tasks, such as math reasoning and coding. Then, for each question in the pool, we evaluate the correctness of each LLM agent’s response, which serves as prior knowledge for the subsequent selection process.

\subsubsection{Pair-wise Reward Ranking}

Given the constructed question pool $\mathcal{Q} = \{q_1, q_2, \dots, q_N\}$, we aim to evaluate whether the reward score provided by the reward models is accurate. That is to say, for the same question, correct answers should receive higher scores than incorrect ones. Specifically, given a question $q \in \mathcal{Q}$, let $\mathcal{A}_q = \{(a_1, y_1), (a_2, y_2), \dots, (a_n, y_n)\}$ denote the set of answers provided by $n$ agents, where $a_i$ is the answer generated by the $i$-th agent and $y_i \in \{0,1\}$ indicates whether the answer is correct ($1$) or incorrect ($0$). $\mathcal{A}_q$ is then partition into two subsets:

\begin{equation}
\left\{
\begin{aligned}
\mathcal{A}_q^{\text{pos}} &= \{a_i \mid y_i = 1\}, \\
\mathcal{A}_q^{\text{neg}} &= \{a_j \mid y_j = 0\},
\end{aligned}
\right.
\end{equation}
where $\mathcal{A}_q^{\text{pos}}$ and $\mathcal{A}_q^{\text{neg}}$ represent correct and incorrect responses, respectively. We then construct all possible pairs $(a_i, a_j)$ where $a_i \in \mathcal{A}_q^{\text{pos}}$ and $a_j \in \mathcal{A}_q^{\text{neg}}$. For each pair, we query the reward function $MoR(\cdot)$ to obtain their reward scores, denoted as $r(a_i)=MoR(a_i,q)$ and $r(a_j)=MoR(a_j,q)$. If $r(a_i)>r(a_j)$, we consider this pair to be accurately assessed by the reward model. The pair-wise accuracy of the reward model on question $q$ can then be defined as:
\begin{equation}
\text{Acc}_q = \frac{1}{|\mathcal{P}_q|} \sum_{(a_i, a_j) \in \mathcal{P}_q} I[r(a_i) > r(a_j)],
\end{equation}
where $\mathcal{P}_q$ is the set of all valid answer pairs for $q$, and $I[\cdot]$ is the indicator function. We can then rank the entire reward model pool on a given question q using $\text{Acc}_q$, which serves as the criterion for subsequent reward model selection. Besides, for questions where all agents provide either entirely correct or entirely incorrect answers, we consider them invalid, as the reward model's accuracy cannot be evaluated on such questions. For multiple reward models $\{R_1, R_2, \dots, R_K\}$, the final reward score for a candidate answer is computed as a weighted combination of the individual reward scores from these $K$ models. Specifically, for a given answer $a$, the reward score from multiple agents is defined as:
\begin{equation}
r_{\text{MoR}}(a,q) = \sum_{k=1}^{K} w_k \cdot R_k(a,q),
\end{equation}
where $w_k$ denotes the weight assigned to reward model $R_k$. The choice of weight computation plays a crucial role in the effectiveness of the MoR. In this work, we basically utilize three weighting strategies based on the individual reward model accuracies $\{\alpha_1, \alpha_2, \dots, \alpha_K\}$ obtained by PRR. For Linear weighting, the weight is proportional to the accuracy:
\begin{equation}
w_k = \frac{\alpha_k}{\sum_{j=1}^{K} \alpha_j}.
\end{equation}
For Softmax weighting, we compute the weights via a softmax by:
\begin{equation}
w_k = \frac{\exp(\alpha_k/\tau)}{\sum_{j=1}^{K} \exp(\alpha_j/\tau)},
\end{equation}
where $\tau > 0$ is a temperature parameter. For naive sum, all reward models are treated equally and no weighting is applied. This corresponds to setting $w_k = 1$ for all $k$.

\subsubsection{Prior Reward Model Ensemble Selection}

Given a question as $q$, a pre-trained embedding model is utilized to embed it into a $d$-dimensional semantic space, resulting in vector $\mathbf{e}_q \in \mathbb{R}^d$. Similarly, the question pool $\mathcal{Q} = \{q_1, \dots, q_N\}$ can be embedded into a matrix $\mathbf{E} \in \mathbb{R}^{N \times d}$, where each row $\mathbf{e}_i$ is the embedding of $q_i$. We then compute the cosine similarity vector $\mathbf{s} \in \mathbb{R}^N$ by $\mathbf{s} =e_q \cdot \mathbf{E}^T$. We select the top-$k$ questions with the highest similarity scores, forming index set $\mathcal{I}_{\text{top}} \subset \{1, \dots, N\}$. For each reward model or combination $R$, we retrieve its pair-wise accuracy vector $\boldsymbol{\alpha}^m \in \mathbb{R}^N$ over the top-k question set. Using the selected indices $\mathcal{I}_{\text{top}}$, we compute a final score by weighted dot product:
\[
\text{Score}_q^m = \sum_{i \in \mathcal{I}_{\text{top}}} s_i \cdot \alpha^m_i.
\]
The final reward model(s) selected for $q$ is:
\[
R_q^* = \arg\max_m \text{Score}_q^m.
\]
We then use the selected reward model(s) for greedy search.

\section{Experiment}

In this section, we first analyze exploratory experiments among different CTTS paradigms. Then we present a comprehensive comparison between our CTTS-MM and existing methods across seven benchmark datasets. Finally, we perform a series of analytical and ablation studies to further investigate the effectiveness of our approach.

\subsection{Experimental Setting}
\label{sec:exp_setting}
\paragraph{Datasets.}
To ensure comprehensive evaluation across diverse capabilities, seven multi-domain datasets across four representative task types are utilized: (1) mathematical reasoning (MATH~\cite{math500}, AIME2024~\cite{AIME2024}); (2) complex knowledge-based reasoning (GPQA~\cite{rein2024gpqa}); (3) instruction-following tasks (IFEval~\cite{ifeval}); (4) code generation (MBPP~\cite{austin2021mbpp}, LiveCodeBench~\cite{jain2024livecodebench}, HumanEval~\cite{humaneval}). All datasets except HumanEval are partitioned into non-overlapping validation and test subsets. Validation portions are utilized to construct the question pool. Details are provided in Section~\ref{dataset} of our Appendix.
\begin{table*}[!th]
\vspace{-6pt}
\small
\setlength{\tabcolsep}{3pt} 
    \centering
    \caption{Comparison results of different TTS paradigms. MR$^*$ means utilizing our proposed MoR to select reward models. Multi-agent$^*$ means utilizing ten chosen LLMs.}
\begin{adjustbox}{max width=\textwidth}
    \begin{tabular}{lllcccccccc}
        \toprule
        
         \textbf{Setting} &  \textbf{Model} & \textbf{Reward Model} & \textbf{Weight Method} & \textbf{MBPP} & \textbf{MATH-500} & \textbf{GPQA} & \textbf{Avg.} \\ 
         
        \midrule 
        
        & Qwen2.5-32B-Instruct & - & - & 76.00 & 75.60 & 40.91 & 64.17 \\
        
        & Qwen2.5-72b-Instruct & - & - & 75.80 & 78.80 & 45.45 & 66.68 \\
        \multicolumn{1}{l}{\multirow{-3}{2cm}{Single Agent}}
        & Llama-3.3-Nemotron-Super-49B-v1 & - & - & 65.40 & 75.20 & 48.48 & 63.03 \\
        
        \midrule

        & Qwen2.5-32B-Instruct & AceCodeRM-32B & - & 77.40 & 78.2 & 47.47 & 67.69 \\
          & Qwen2.5-32B-Instruct & Qwen2.5-Math-RM-72B & - & 77.00 & 78.80 & 46.46 & 67.42 \\
        & Qwen2.5-72b-Instruct & AceCodeRM-32B & - & 76.60 & 80.20 & 51.01 & 69.27 \\
        & Qwen2.5-72b-Instruct & Qwen2.5-Math-RM-72B & - & 76.00 & 80.80 & 50.51 & 69.10 \\
        
        &Llama-3.3-Nemotron-Super-49B-v1 & AceCodeRM-32B & - & 66.40 & 76.00 & 50.80 & 64.40 \\
        \multicolumn{1}{l}{\multirow{-6}{2cm}{SA-SR}}
        & Llama-3.3-Nemotron-Super-49B-v1 & Qwen2.5-Math-RM-72B & - & 65.80 & 76.80 & 50.00 & 64.20 \\

        \midrule   
        
        & Qwen2.5-32B-Instruct & AceCodeRM-32B+Qwen2.5-Math-RM-72B & softmax & 76.6 & 78.2 & 48.48 & 67.76 \\
        & Qwen2.5-32b-Instruct & MR$^*$ & - & 78.00 & 79.4 & 51.01 & 69.47 \\
        
        & Qwen2.5-72B-Instruct & AceCodeRM-32B+Qwen2.5-Math-RM-72B & softmax & 76.8 & 80.20 & 51.51 & 69.50 \\
        & Qwen2.5-72b-Instruct & MR$^*$ & - & 77.20 & 81.4 & 53.53 & 70.71 \\

        &Llama-3.3-Nemotron-Super-49B-v1 & AceCodeRM-32B+Qwen2.5-Math-RM-72B & softmax & 66.20 & 76.60 & 51.52 & 64.77 \\
        \multicolumn{1}{l}{\multirow{-6}{2cm}{SA-MR}}
        & Llama-3.3-Nemotron-Super-49B-v1 & MR$^*$ & - & 66.80 & 76.80 & 54.55 & 66.05 \\

        \midrule   

        & Multi-agent$^*$ & Skywork-Reward-V2-Llama-3.1-8B-40M & - & 77.00 & \cellcolor{toptwoRed}91.20 & \cellcolor{toptwoRed}61.11 & 75.97 \\
        
        & Multi-agent$^*$ & Qwen2.5-Math-RM-72B & - & \cellcolor{toptwoRed}80.6 & \cellcolor{toponeRed}91.8 & \cellcolor{toptwoRed}61.11 & \cellcolor{toptwoRed}77.84 \\
        
        \multicolumn{1}{l}{\multirow{-3}{2cm}{MA-SR}}
        
        & Multi-agent$^*$ & AceCodeRM-32B & - & \cellcolor{toponeRed}82.2 & 90.8 & \cellcolor{toponeRed}61.62 & \cellcolor{toponeRed}78.21 \\

        \midrule   

        \multicolumn{1}{l}{MA-MR (Proposed CTTS-MM)}& Multi-agent$^*$ & MR$^*$ & - & \textbf{83.20} & \textbf{93.00} & \textbf{64.14} & \textbf{80.11} \\

        \bottomrule
    \end{tabular}
    \end{adjustbox}
    \vspace{-15pt}

    \label{table:paradigm_main}
\end{table*}

\begin{table*}[th!]

  \centering
      \caption{Main Results of CTTS-MM compared with leading LLMs and other related methods on seven mainstream benchmarks. }
\begin{adjustbox}{max width=\textwidth}
    \begin{tabular}{lcccccccc}
      \toprule
      \textbf{Model} & \textbf{AIME-2024} & \textbf{MATH-500} & \textbf{MBPP} & \textbf{LiveCodeBench} & \textbf{GPQA-Diamond} & \textbf{Human-eval} & \textbf{IFEval}  & \textbf{Avg} \\
      \midrule
      
      \rowcolor{gray!100} \multicolumn{9}{c}{\textit{Open-source LLMs}} \\

      Qwen-2.5-72B-Instruct & 16.70 & 78.80 & 75.80 & 26.10 & 45.45 & 78.66 & 86.30 & 58.26 \\
      
      DeepSeek-R1-Distill-Llama-70B & 60.00 & 82.80 & 76.40 & 40.70 & 60.10 & 92.07 & 80.30 & 70.34 \\
      Llama-3.3-Nemotron-Super-49B-v1 & 16.70 & 75.20 & 65.40 & 28.00 & 48.48 & 84.76 & 82.70 & 57.32 \\
      QwQ-32B & 46.70 & 87.80 & 81.80 & 38.60 & 57.07 & 92.07 & 81.70  & 69.39 \\
      
      InternLM2.5-20B-Chat & 3.30 & 55.20 & 55.00 & 14.90 & 34.85 & 69.51 & 64.70 & 42.49 \\
      
      Gemma-3-27b-it & 30.00 & 84.00 & 70.40 & 27.70 & 50.51 & 86.59 & 81.00  & 61.46 \\
      
      Qwen2.5-32b-Instruct & 20.00 & 75.60 & 76.00 & 24.00 & 40.91 & 77.44 & 78.70  & 56.09 \\
      
      TeleChat2-35B-32K & 10.00 & 70.00 & 70.00 & 19.50 & 33.33 & 73.17 & 82.00  & 51.14 \\
        EXAONE-Deep-32B & 33.30 & 84.38 & 72.80 & 31.60 & 58.59 & 93.90 & 76.30  & 64.41 \\
     
      GLM-Z1-32B-0414 & \cellcolor{toptwoRed}66.70 & 90.00 & 74.40 & 44.40 & 59.60 & \cellcolor{toptwoRed}96.34 & 83.00 & 73.49 \\
      Llama-3.3-70B-Instruct & 30.00 & 73.00 & 70.40 & 30.10 & 46.97 & 84.15 & \cellcolor{toptwoRed}90.00 & 60.66 \\
      
 Qwen3-32B & 53.30 & 88.00 & 50.60 & 33.40 & \cellcolor{toponeRed}65.15 & 90.85 & 83.70 & 66.43 \\
      Qwen2.5-Coder-32B-Instruct & 16.70 & 73.60 & 78.00 & 27.70 & 41.92 & 87.80 & 80.30 & 58.00 \\
      
      HuatuoGPT-o1-72B & 16.70 & 73.00 & 78.00 & 27.40 & 50.00 & 85.37 & 74.00 & 57.78 \\

      DeepSeek-R1-Distill-Qwen-32B & 56.70 & 85.60 & 81.00 & 44.70 & 60.10 & 95.73 & 73.70 & 71.08 \\

      \rowcolor{gray!100} \multicolumn{9}{c}{\textit{Proprietary LLMs}} \\
          GPT-4.1 (2025-04-14) & 50.00 & 85.80 & 79.20 & 42.20 & 67.17 & 92.07 & 86.00  & 71.78 \\
      
      Claude-3.7-Sonnet (2025-02-19) & 26.70 & 73.20 & 75.40 & 41.30 & 63.64 & 90.85 & 88.00 & 65.58 \\
      
      GPT-4o (2024-08-06) & 10.00 & 74.60 & 74.20 & 29.80 & 52.53 & 85.36 & 82.30  & 58.40 \\
      
      Claude-3.5-Sonnet (2024-06-20) & 16.70 & 74.20 & 75.80 & 34.30 & 61.62 & 89.63 & 80.30  & 61.79 \\

      \rowcolor{gray!100} \multicolumn{9}{c}{\textit{Related Methods}} \\

        Majority Voting~\cite{chen2024are} &  56.67   &     90.20      &   80.40  &     34.65          &    26.26         &  89.63         &   80.67       &  65.50  \\
        Multi-Agent Verification~\cite{lifshitz2025multiagent} &  63.33   &     76.30      &   74.60  &        42.55       &    59.00         &  92.00         &   83.00       &  70.11  \\
      
      Symbolic-MoE~\cite{Chen2025SymbolicMA} &  50.00   &  90.40  &  \cellcolor{toptwoRed}82.60   &  43.16 &  62.63    &           92.07  & 89.00   & 72.82\\
      
      MoA~\cite{wang2025mixtureofagents} &  53.33   &    87.80       &  82.00   &    40.12           &    58.80         &   90.85       &    89.33         &  71.75  \\

      Self Consistency~\cite{chen2024universal} &  \cellcolor{toponeRed}70.00  &   \cellcolor{toptwoRed}91.40   &   82.40   &   30.47  & \cellcolor{toponeRed}65.15  & 90.39   &  68.33   &  71.16 \\

        Best of N~\cite{bestofn} &  66.70  &  90.8   &   75.00  &   \cellcolor{toptwoRed}44.98  & 60.61  &  \cellcolor{toptwoRed}96.34  &  83.66   &  \cellcolor{toptwoRed}74.01 \\
    
      \rowcolor{gray!100} \multicolumn{9}{c}{\textit{Ours v.s. Strong Baselines}} \\
      \textbf{CTTS-MM(ours)} & \cellcolor{toponeRed}70.00 & \cellcolor{toponeRed}93.00 & \cellcolor{toponeRed}83.20 & \cellcolor{toponeRed}52.28 & 
      \cellcolor{toptwoRed}64.14 & \cellcolor{toponeRed}97.56 & \cellcolor{toponeRed}91.67 & \cellcolor{toponeRed}78.84 \\
      
      \textit{- v.s. GLM-Z1-32B-0414} & \positive{3.30} & \positive{3.00} & \positive{8.80} & \positive{7.88} & \positive{4.54} & \positive{1.22} & \positive{8.67}  & \positive{5.34} \\
      
      \textit{- v.s. GPT-4.1} & \positive{20.00} & \positive{7.20} & \positive{4.00} & \positive{10.08} & \negative{3.03} & \positive{5.49} & \positive{5.67}  & \positive{7.06} \\
      
      \textit{- v.s. Best of N} & \positive{3.30} & \positive{2.20} & \positive{8.20} & \positive{7.29} & \positive{3.53} & \positive{1.22} & \positive{8.01} & \positive{4.82} \\
      
            \bottomrule
    \end{tabular}
  \end{adjustbox}
    \vspace{-13pt}

  \label{tab:main_results_main}  
\end{table*}

\begin{table}[h]
\centering
\caption{Component ablation on four standard datasets.}
\begin{adjustbox}{max width=0.75\textwidth}
\begin{tabular}{ccc|cccc}
\toprule
MoR & ACS   & Residual Aggregation
& MATH-500 & MBPP & AIME & LiveCodeBench \\ \midrule
\xmark     & \xmark     & \xmark         & 90.80    & 80.00  & 56.67 & 40.12   \\
\xmark     & \checkmark & \xmark        & 91.20    & 80.20      & 60.00   & 43.16 \\
\checkmark & \xmark & \xmark         & 91.40    & 80.20   & 63.33 & 43.77 \\
\xmark     & \checkmark & \checkmark       & 91.80     & 80.60   & 66.67 & 44.38 \\

\checkmark & \checkmark & \xmark     & 92.40    & 83.00  & 70.00 & 51.67 \\
\checkmark & \checkmark & \checkmark  & 93.00    & 83.20   & 70.00 & 52.28 \\ \bottomrule
\end{tabular}
\end{adjustbox}

\vspace{-10pt}
\label{tab:ablation}

\end{table}

\begin{figure}[t]
  \centering
  \includegraphics[width=0.75\columnwidth]{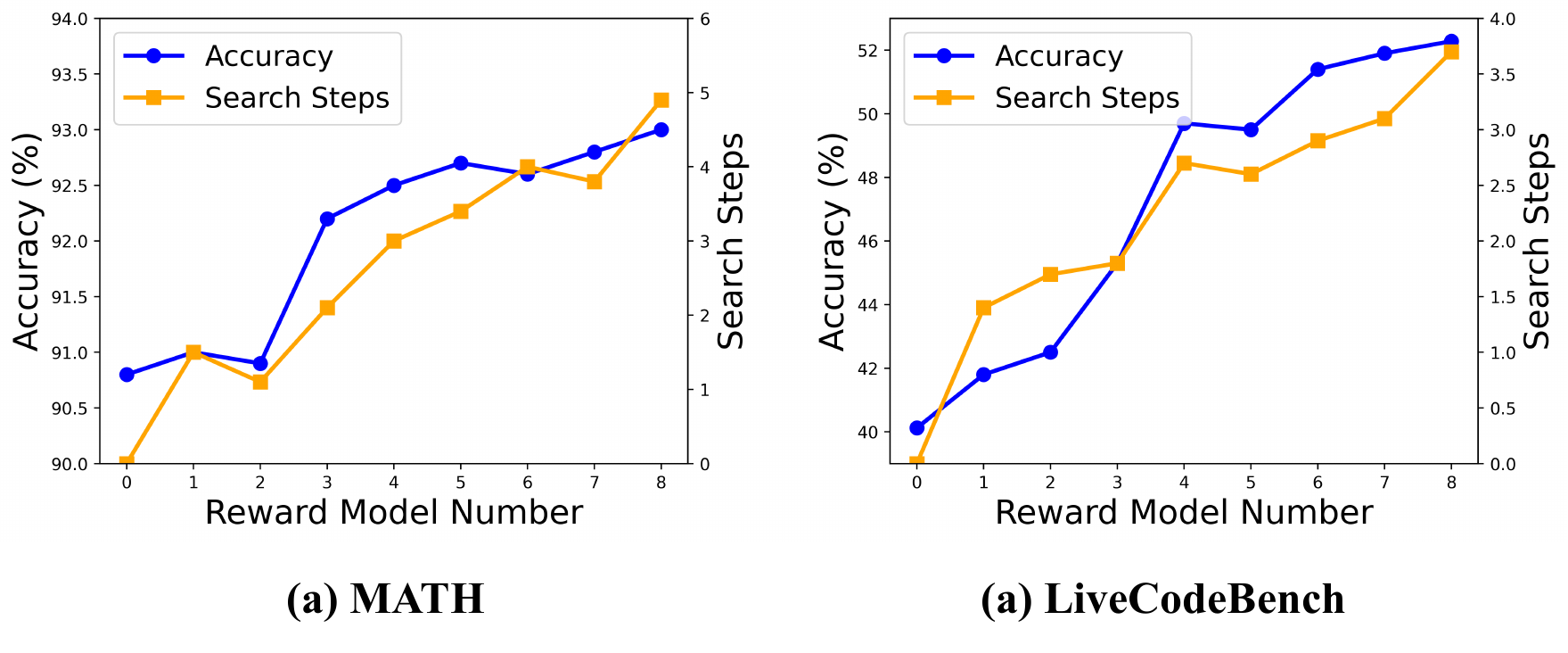}
    \vspace{-10pt}
  \caption{The curve of scaling capability of Reward Models.}
  \vspace{-15pt}

  \label{fig:rm_acc}
\end{figure}

\paragraph{LLM agents and Reward Models.}
For LLM agents used in our experiments, we assemble a set of ten mid-sized open-source LLMs (ranging from 20B to 72B parameters) from diverse architecture families. For reward models, we select eight off-the-shelf models, covering specialized domains like math and coding. Additional details are provided in the Appendix.

\subsection{Analysis on Different TTS Paradigms}

To thoroughly investigate CTTS and STTS paradigms, we conduct exploratory experiments measuring performance variation among different paradigms. As demonstrated in Table~\ref{table:paradigm_main}, CTTS paradigms outperform both STTS and single-model baselines under most settings. For example, under MA-MR setting, our method achieves an average improvement of 10.84\% (80.11\% v.s. 69.27\%) over the best performance in the SA-SR setting, with gains of 5.8\% (83.20\% v.s. 77.40\%) on MBPP, 12.20\% (93.00\% v.s. 80.80\%) on MATH, and 13.13\% (64.14 vs. 51.01) on GPQA, respectively. Under MA-SR setting, the best performance shows an average improvement of 8.94\% over SA-SR while SA-MR yields an average gain of 1.44\%. The results indicate the effectiveness of our proposed CTTS paradigm. In particular, our MA-MR based framework achieves superior improvements. Besides, we observe that under the SA-MR setting, a fixed combination of reward models may lead to performance degradation compared to SA-SR. This indicates that naively combining multiple reward models is unlikely to yield improvements and can even bring performance dropping. Such results are expected since most reward models are domain-specific. Fixed combination cannot guarantee consistent gains across all problems from different domains, which underscores the necessity of our proposed MoR for adaptive reward model selection. Comparative results between MA-MR and MA-SR also demonstrate that multi-RM achieves superior generalization compared to domain-specific reward models. Refer to our Appendix for more results.

\vspace{-10pt}
\subsection{Main results}

As demonstrated in Table~\ref{tab:main_results_main}, our proposed CTTS-MM demonstrates significant improvements across seven diverse benchmarks. Through comprehensive comparisons with (1) fifteen representative open-source models (2) four leading Proprietary models and (3) five existing collaboration methods, our approach demonstrates consistent and substantial improvements across all evaluation dimensions. Our framework achieves 78.84\% average accuracy on seven benchmarks. Compared to existing collaboration approaches, CTTS-MM outperforms Majority Voting~\cite{chen2024are} by +13.34\%, MAV~\cite{lifshitz2025multiagent} by +8.73\%, Symbolic-MoE~\cite{Chen2025SymbolicMA} by +6.02\%, MoA~\cite{wang2025mixtureofagents} by 7.09\%, Self Consistency~\cite{chen2024universal} by 7.68\% and Best of N~\cite{bestofn} by 4.83\%. Remarkably, our approach demonstrates superior performance compared to strong baselines from open-source LLM agents, proprietary LLM agents, and related multi-agent methods. Specifically, on average accuracy, CTTS-MM surpasses the best-performing open-source LLM GLM-Z1-32B by 5.34\%, proprietary LLM GPT-4.1 by 7.06\%, and STTS-based method Best-of-N by 4.82\%. These results demonstrate that our CTTS-MM can effectively leverage the complementary advantages of multiple agents and multiple reward models, leading to a superior performance increase. This further validates the substantial potential of the CTTS paradigm, particularly highlighting the MA-MR framework's robust capability in multiple domains. Refer to our Appendix for more results.

\begin{table}[!t]
\centering
\caption{Comparative results of average inference time on MATH-500 and LiveCodeBench.}
\begin{adjustbox}{max width=0.75\textwidth}
\begin{tabular}{l|cc|cc}
\toprule
\multirow{2}{*}{Method} & \multicolumn{2}{c}{MATH-500}       & \multicolumn{2}{c}{LiveCodeBench}           \\ \cmidrule{2-5} 
                        & Inference Time (s) & Accuracy (\%) & Inference Time (s) & Accuracy (\%)  \\ \midrule
Multi-Agent Verification       & 22.45         & 76.30      & 24.52       & 42.55     \\ 
Symbolic-MoE       & 15.05         & 90.40      & 17.76       & 43.16     \\
MoA       & 19.54         & 87.80      & 21.97      & 40.12     \\
Best of N       & 13.12         & 90.80      & 16.17       & 44.98     \\
\midrule
CTTS-MM (Ours)     &  20.14                 & 93.00             &   22.04                & 52.28      
\\ \bottomrule
\end{tabular}
\end{adjustbox}
\label{tab:inference_time}
\end{table}

\vspace{-10pt}

\subsection{Ablation Study}
We perform a comprehensive component-wise ablation study on four standard benchmarks to quantify the contribution of each component in our CTTS-MM framework. Note that Residual Aggregation can only be applied when Agent Collaboration Search (ACS) is utilized. As illustrated in Table~\ref{tab:ablation}, the baseline obtains 90.08\% accuracy on MATH. Utilizing ACS and MoR improves performance by 0.4\% and 0.6\%, respectively, reaching 92.40\% when combined. Further gains come from Residual Aggregation, which contributes an additional 0.6\%. Similar improvements are observed on the other three benchmarks, indicating the effectiveness of each component in CTTS-MM. 

\subsection{Analysis on Scaling Capability of RMs}
To investigate the scalability of MoR, we conduct experiments measuring performance improvements with increasing numbers of reward models. As shown in Figure~\ref{fig:rm_acc}, the performance of CTTS-MM consistently improves with increasing number of reward models on both MATH and LiveCodeBench. For instance, on LiveCodeBench, CTTS-MM achieves an accuracy of approximately 41.8\% with a single reward model. When the number of reward models increases to four, the accuracy improves to nearly 50\%, and ultimately reaches 52.28\% with all eight reward models. Moreover, we observe that the search step also gradually increases with more reward models. It indicates that with the increase of reward models, our Mixture of Reward Model can enhance the entire model pool to extract cross-domain information, leading to more accurate reward scores for question-answer pairs during search. It guides the search process in a more optimal direction. In contrast, when the RM pool is limited in size, its robustness is weaker, resulting in inaccurate reward scores. This can mislead the search direction, potentially causing early stop and local optimal solutions. 

\subsection{Analysis on Time and Cost Efficiency}
Although we primarily focus on exploring the optimal paradigm under collective test-time scaling, we further investigate the inference time and cost efficiency of our CTTS-MM in comparison to other related methods. For time efficiency, we report average inference time per response on MATH-500 and LiveCodeBench. As shown in Table~\ref{tab:inference_time}, CTTS-MM achieves a significant improvement in accuracy with bearable inference time compared to other methods. For cost efficiency, we plot results of performance against inference cost. Refer to Section~\ref{cost} of our Appendix for the detailed figure.

\section{Conclusion}
In this manuscript, we first explore Collective Test-Time Scaling (CTTS). We propose and investigate three CTTS paradigms: SA-MR, MA-SR and MA-MR. Experiments demonstrate that CTTS outperforms previous Single TTS paradigms, while the MA-MR variant consistently achieves superior performance. Based on it, we further propose a CTTS framework called CTTS-MM. To search for optimal agent ensembles, we propose Agent Collaboration Search approach. For adaptively selecting multiple reward models, Prior Reward Models Ensemble Selection is proposed. Experiments on seven benchmarks verify the superiority of CTTS-MM, revealing the strong potential of CTTS.

\bibliography{iclr2026_conference}

\begin{thebibliography}{56}
\providecommand{\natexlab}[1]{#1}
\providecommand{\url}[1]{\texttt{#1}}
\expandafter\ifx\csname urlstyle\endcsname\relax
  \providecommand{\doi}[1]{doi: #1}\else
  \providecommand{\doi}{doi: \begingroup \urlstyle{rm}\Url}\fi

\bibitem[Austin et~al.(2021)Austin, Odena, Nye, Bosma, Michalewski, Dohan, Jiang, Cai, Terry, Le, et~al.]{austin2021mbpp}
Jacob Austin, Augustus Odena, Maxwell Nye, Maarten Bosma, Henryk Michalewski, David Dohan, Ellen Jiang, Carrie Cai, Michael Terry, Quoc Le, et~al.
\newblock Program synthesis with large language models.
\newblock \emph{arXiv preprint arXiv:2108.07732}, 2021.

\bibitem[Bercovich et~al.(2025)Bercovich, Levy, Golan, Dabbah, El-Yaniv, Puny, Galil, Moshe, Ronen, Nabwani, et~al.]{bercovich2025llama}
Akhiad Bercovich, Itay Levy, Izik Golan, Mohammad Dabbah, Ran El-Yaniv, Omri Puny, Ido Galil, Zach Moshe, Tomer Ronen, Najeeb Nabwani, et~al.
\newblock Llama-nemotron: Efficient reasoning models.
\newblock \emph{arXiv preprint arXiv:2505.00949}, 2025.

\bibitem[Brown et~al.(2024)Brown, Juravsky, Ehrlich, Clark, Le, R'e, and Mirhoseini]{monkey}
Bradley Brown, Jordan Juravsky, Ryan Ehrlich, Ronald Clark, Quoc~V. Le, Christopher R'e, and Azalia Mirhoseini.
\newblock Large language monkeys: Scaling inference compute with repeated sampling.
\newblock \emph{ArXiv}, abs/2407.21787, 2024.

\bibitem[Brown et~al.(2020)Brown, Mann, Ryder, Subbiah, Kaplan, Dhariwal, Neelakantan, Shyam, Sastry, Askell, Agarwal, Herbert-Voss, Krueger, Henighan, Child, Ramesh, Ziegler, Wu, Winter, Hesse, Chen, Sigler, teusz Litwin, Gray, Chess, Clark, Berner, McCandlish, Radford, Sutskever, and Amodei]{Brown2020LanguageMA}
Tom~B. Brown, Benjamin Mann, Nick Ryder, Melanie Subbiah, Jared Kaplan, Prafulla Dhariwal, Arvind Neelakantan, Pranav Shyam, Girish Sastry, Amanda Askell, Sandhini Agarwal, Ariel Herbert-Voss, Gretchen Krueger, T.~J. Henighan, Rewon Child, Aditya Ramesh, Daniel~M. Ziegler, Jeff Wu, Clemens Winter, Christopher Hesse, Mark Chen, Eric Sigler, Ma~teusz Litwin, Scott Gray, Benjamin Chess, Jack Clark, Christopher Berner, Sam McCandlish, Alec Radford, Ilya Sutskever, and Dario Amodei.
\newblock Language models are few-shot learners.
\newblock \emph{ArXiv}, abs/2005.14165, 2020.

\bibitem[Cai et~al.(2024)Cai, Cao, Chen, Chen, Chen, Chen, Chen, Chen, Chen, Chu, Dong, Duan, Fan, Fei, Gao, Ge, Gu, Gu, Gui, Guo, Guo, He, Hu, Huang, Jiang, Jiao, Jin, Lei, Li, Li, Li, Li, Li, Li, Liu, Liu, Hong, Liu, Liu, Liu, Lv, Lv, Lv, Ma, Ma, Ma, Ning, Ouyang, Qiu, Qu, Shang, Shao, Song, Song, Sui, Sun, Sun, Tang, Wang, Wang, Wang, Wang, Wang, Wang, Wang, Wei, Weng, Wu, Xiong, Xu, Xu, Yan, Yan, Yang, Ye, Ying, Yu, Yu, Zang, Zhang, Zhang, Zhang, Zhang, Zhang, Zhang, Zhang, Zhang, Zhang, Zhang, Zhang, Zhao, Zhao, Zhao, Zhou, Zhou, Zhuo, Zou, Qiu, Qiao, and Lin]{cai2024internlm2}
Zheng Cai, Maosong Cao, Haojiong Chen, Kai Chen, Keyu Chen, Xin Chen, Xun Chen, Zehui Chen, Zhi Chen, Pei Chu, Xiaoyi Dong, Haodong Duan, Qi~Fan, Zhaoye Fei, Yang Gao, Jiaye Ge, Chenya Gu, Yuzhe Gu, Tao Gui, Aijia Guo, Qipeng Guo, Conghui He, Yingfan Hu, Ting Huang, Tao Jiang, Penglong Jiao, Zhenjiang Jin, Zhikai Lei, Jiaxing Li, Jingwen Li, Linyang Li, Shuaibin Li, Wei Li, Yining Li, Hongwei Liu, Jiangning Liu, Jiawei Hong, Kaiwen Liu, Kuikun Liu, Xiaoran Liu, Chengqi Lv, Haijun Lv, Kai Lv, Li~Ma, Runyuan Ma, Zerun Ma, Wenchang Ning, Linke Ouyang, Jiantao Qiu, Yuan Qu, Fukai Shang, Yunfan Shao, Demin Song, Zifan Song, Zhihao Sui, Peng Sun, Yu~Sun, Huanze Tang, Bin Wang, Guoteng Wang, Jiaqi Wang, Jiayu Wang, Rui Wang, Yudong Wang, Ziyi Wang, Xingjian Wei, Qizhen Weng, Fan Wu, Yingtong Xiong, Chao Xu, Ruiliang Xu, Hang Yan, Yirong Yan, Xiaogui Yang, Haochen Ye, Huaiyuan Ying, Jia Yu, Jing Yu, Yuhang Zang, Chuyu Zhang, Li~Zhang, Pan Zhang, Peng Zhang, Ruijie Zhang, Shuo Zhang, Songyang Zhang, Wenjian Zhang,
  Wenwei Zhang, Xingcheng Zhang, Xinyue Zhang, Hui Zhao, Qian Zhao, Xiaomeng Zhao, Fengzhe Zhou, Zaida Zhou, Jingming Zhuo, Yicheng Zou, Xipeng Qiu, Yu~Qiao, and Dahua Lin.
\newblock Internlm2 technical report, 2024.

\bibitem[Chen et~al.(2024{\natexlab{a}})Chen, Lin, Han, and Sun]{benchret}
Jiawei Chen, Hongyu Lin, Xianpei Han, and Le~Sun.
\newblock Benchmarking large language models in retrieval-augmented generation.
\newblock In \emph{Proceedings of the Thirty-Eighth AAAI Conference on Artificial Intelligence and Thirty-Sixth Conference on Innovative Applications of Artificial Intelligence and Fourteenth Symposium on Educational Advances in Artificial Intelligence}, AAAI'24/IAAI'24/EAAI'24. AAAI Press, 2024{\natexlab{a}}.
\newblock ISBN 978-1-57735-887-9.

\bibitem[Chen et~al.(2024{\natexlab{b}})Chen, Cai, Ji, Wang, Liu, Wang, Hou, and Wang]{chen2024huatuogpto1medicalcomplexreasoning}
Junying Chen, Zhenyang Cai, Ke~Ji, Xidong Wang, Wanlong Liu, Rongsheng Wang, Jianye Hou, and Benyou Wang.
\newblock Huatuogpt-o1, towards medical complex reasoning with llms, 2024{\natexlab{b}}.

\bibitem[Chen et~al.(2025)Chen, Yun, Stengel-Eskin, Chen, and Bansal]{Chen2025SymbolicMA}
Justin Chih-Yao Chen, Sukwon Yun, Elias Stengel-Eskin, Tianlong Chen, and Mohit Bansal.
\newblock Symbolic mixture-of-experts: Adaptive skill-based routing for heterogeneous reasoning.
\newblock \emph{ArXiv}, abs/2503.05641, 2025.

\bibitem[Chen et~al.(2024{\natexlab{c}})Chen, Davis, Hanin, Bailis, Stoica, Zaharia, and Zou]{chen2024are}
Lingjiao Chen, Jared~Quincy Davis, Boris Hanin, Peter Bailis, Ion Stoica, Matei Zaharia, and James Zou.
\newblock Are more {LLM} calls all you need? towards the scaling properties of compound {AI} systems.
\newblock In \emph{The Thirty-eighth Annual Conference on Neural Information Processing Systems}, 2024{\natexlab{c}}.

\bibitem[Chen et~al.(2024{\natexlab{d}})Chen, Zaharia, and Zou]{chen2024frugalgpt}
Lingjiao Chen, Matei Zaharia, and James Zou.
\newblock Frugal{GPT}: How to use large language models while reducing cost and improving performance.
\newblock \emph{Transactions on Machine Learning Research}, 2024{\natexlab{d}}.
\newblock ISSN 2835-8856.

\bibitem[Chen(2025)]{chen2025LDLRewardGemma}
Shikai Chen.
\newblock Ldl-reward-gemma-2-27b-v0.1, 2025.
\newblock Accessed: 2025-02-15.

\bibitem[Chen et~al.(2024{\natexlab{e}})Chen, Aksitov, Alon, Ren, Xiao, Yin, Prakash, Sutton, Wang, and Zhou]{chen2024universal}
Xinyun Chen, Renat Aksitov, Uri Alon, Jie Ren, Kefan Xiao, Pengcheng Yin, Sushant Prakash, Charles Sutton, Xuezhi Wang, and Denny Zhou.
\newblock Universal self-consistency for large language models.
\newblock In \emph{ICML 2024 Workshop on In-Context Learning}, 2024{\natexlab{e}}.

\bibitem[DeepSeek-AI(2025)]{deepseekai2025}
DeepSeek-AI.
\newblock Deepseek-r1: Incentivizing reasoning capability in llms via reinforcement learning, 2025.

\bibitem[DeepSeek-AI \& et~al.(2025)DeepSeek-AI and et~al.]{DeepSeekAI2025DeepSeekR1IR}
DeepSeek-AI and Daya~Guo et~al.
\newblock Deepseek-r1: Incentivizing reasoning capability in llms via reinforcement learning.
\newblock \emph{ArXiv}, abs/2501.12948, 2025.

\bibitem[Dorka(2024)]{qrmgemma}
Nicolai Dorka.
\newblock Quantile regression for distributional reward models in rlhf.
\newblock \emph{ArXiv}, abs/2409.10164, 2024.

\bibitem[Du et~al.(2024)Du, Li, Torralba, Tenenbaum, and Mordatch]{debate}
Yilun Du, Shuang Li, Antonio Torralba, Joshua~B. Tenenbaum, and Igor Mordatch.
\newblock Improving factuality and reasoning in language models through multiagent debate.
\newblock In \emph{Proceedings of the 41st International Conference on Machine Learning}, ICML'24. JMLR.org, 2024.

\bibitem[GLM et~al.(2024)GLM, Zeng, Xu, Wang, Zhang, Yin, Rojas, Feng, Zhao, Lai, Yu, Wang, Sun, Zhang, Cheng, Gui, Tang, Zhang, Li, Zhao, Wu, Zhong, Liu, Huang, Zhang, Zheng, Lu, Duan, Zhang, Cao, Yang, Tam, Zhao, Liu, Xia, Zhang, Gu, Lv, Liu, Liu, Yang, Song, Zhang, An, Xu, Niu, Yang, Li, Bai, Dong, Qi, Wang, Yang, Du, Hou, and Wang]{glm2024chatglm}
Team GLM, Aohan Zeng, Bin Xu, Bowen Wang, Chenhui Zhang, Da~Yin, Diego Rojas, Guanyu Feng, Hanlin Zhao, Hanyu Lai, Hao Yu, Hongning Wang, Jiadai Sun, Jiajie Zhang, Jiale Cheng, Jiayi Gui, Jie Tang, Jing Zhang, Juanzi Li, Lei Zhao, Lindong Wu, Lucen Zhong, Mingdao Liu, Minlie Huang, Peng Zhang, Qinkai Zheng, Rui Lu, Shuaiqi Duan, Shudan Zhang, Shulin Cao, Shuxun Yang, Weng~Lam Tam, Wenyi Zhao, Xiao Liu, Xiao Xia, Xiaohan Zhang, Xiaotao Gu, Xin Lv, Xinghan Liu, Xinyi Liu, Xinyue Yang, Xixuan Song, Xunkai Zhang, Yifan An, Yifan Xu, Yilin Niu, Yuantao Yang, Yueyan Li, Yushi Bai, Yuxiao Dong, Zehan Qi, Zhaoyu Wang, Zhen Yang, Zhengxiao Du, Zhenyu Hou, and Zihan Wang.
\newblock Chatglm: A family of large language models from glm-130b to glm-4 all tools, 2024.

\bibitem[Grattafiori et~al.(2024)Grattafiori, Dubey, Jauhri, Pandey, Kadian, Al-Dahle, Letman, Mathur, Schelten, Vaughan, et~al.]{grattafiori2024llama}
Aaron Grattafiori, Abhimanyu Dubey, Abhinav Jauhri, Abhinav Pandey, Abhishek Kadian, Ahmad Al-Dahle, Aiesha Letman, Akhil Mathur, Alan Schelten, Alex Vaughan, et~al.
\newblock The llama 3 herd of models.
\newblock \emph{arXiv preprint arXiv:2407.21783}, 2024.

\bibitem[Gui et~al.(2024)Gui, Garbacea, and Veitch]{gui2024bonbon}
Lin Gui, Cristina Garbacea, and Victor Veitch.
\newblock Bo{NB}on alignment for large language models and the sweetness of best-of-n sampling.
\newblock In \emph{The Thirty-eighth Annual Conference on Neural Information Processing Systems}, 2024.

\bibitem[Hendrycks et~al.(2021)Hendrycks, Burns, Kadavath, Arora, Basart, Tang, Song, and Steinhardt]{math500}
Dan Hendrycks, Collin Burns, Saurav Kadavath, Akul Arora, Steven Basart, Eric Tang, Dawn Song, and Jacob Steinhardt.
\newblock Measuring mathematical problem solving with the math dataset.
\newblock \emph{arXiv preprint arXiv:2103.03874}, 2021.

\bibitem[Hui et~al.(2024)Hui, Yang, Cui, Yang, Liu, Zhang, Liu, Zhang, Yu, Dang, et~al.]{hui2024qwen2}
Binyuan Hui, Jian Yang, Zeyu Cui, Jiaxi Yang, Dayiheng Liu, Lei Zhang, Tianyu Liu, Jiajun Zhang, Bowen Yu, Kai Dang, et~al.
\newblock Qwen2. 5-coder technical report.
\newblock \emph{arXiv preprint arXiv:2409.12186}, 2024.

\bibitem[Jain et~al.(2024)Jain, Han, Gu, Li, Yan, Zhang, Wang, Solar-Lezama, Sen, and Stoica]{jain2024livecodebench}
Naman Jain, King Han, Alex Gu, Wen-Ding Li, Fanjia Yan, Tianjun Zhang, Sida Wang, Armando Solar-Lezama, Koushik Sen, and Ion Stoica.
\newblock Livecodebench: Holistic and contamination free evaluation of large language models for code.
\newblock \emph{arXiv preprint arXiv:2403.07974}, 2024.

\bibitem[junyou li et~al.(2024)junyou li, Zhang, Yu, FU, and Ye]{li2024more}
junyou li, Qin Zhang, Yangbin Yu, QIANG FU, and Deheng Ye.
\newblock More agents is all you need.
\newblock \emph{Transactions on Machine Learning Research}, 2024.
\newblock ISSN 2835-8856.

\bibitem[Kim et~al.(2024)Kim, Chanyeol, Kim, and Lee]{kim2024linq}
Jihoon Kwon Sangmo Gu~Yejin Kim, Minkyung Cho Jy-yong~Sohn Chanyeol, Choi~Junseong Kim, and Seolhwa Lee.
\newblock Linq-embed-mistral: Elevating text retrieval with improved gpt data through task-specific control and quality refinement. linq ai research blog, 2024.

\bibitem[Kwon et~al.(2023)Kwon, Li, Zhuang, Sheng, Zheng, Yu, Gonzalez, Zhang, and Stoica]{kwon2023efficient}
Woosuk Kwon, Zhuohan Li, Siyuan Zhuang, Ying Sheng, Lianmin Zheng, Cody~Hao Yu, Joseph~E. Gonzalez, Hao Zhang, and Ion Stoica.
\newblock Efficient memory management for large language model serving with pagedattention.
\newblock In \emph{Proceedings of the ACM SIGOPS 29th Symposium on Operating Systems Principles}, 2023.

\bibitem[Lewis et~al.(2020)Lewis, Perez, Piktus, Petroni, Karpukhin, Goyal, K\"{u}ttler, Lewis, Yih, Rockt\"{a}schel, Riedel, and Kiela]{retrieval}
Patrick Lewis, Ethan Perez, Aleksandra Piktus, Fabio Petroni, Vladimir Karpukhin, Naman Goyal, Heinrich K\"{u}ttler, Mike Lewis, Wen-tau Yih, Tim Rockt\"{a}schel, Sebastian Riedel, and Douwe Kiela.
\newblock Retrieval-augmented generation for knowledge-intensive nlp tasks.
\newblock In \emph{Proceedings of the 34th International Conference on Neural Information Processing Systems}, NIPS '20, Red Hook, NY, USA, 2020. Curran Associates Inc.
\newblock ISBN 9781713829546.

\bibitem[{LG AI Research}(2025)]{exaone-deep}
{LG AI Research}.
\newblock Exaone deep: Reasoning enhanced language models.
\newblock \emph{arXiv preprint arXiv:2503.12524}, 2025.

\bibitem[Lifshitz et~al.(2025)Lifshitz, McIlraith, and Du]{lifshitz2025multiagent}
Shalev Lifshitz, Sheila~A. McIlraith, and Yilun Du.
\newblock Multi-agent verification: Scaling test-time compute with goal verifiers.
\newblock In \emph{ICLR 2025 Workshop on Modularity for Collaborative, Decentralized, and Continual Deep Learning}, 2025.

\bibitem[Liu et~al.(2024)Liu, Zeng, Liu, Yan, He, Wang, Yan, Liu, and Zhou]{Liu2024SkyworkReward}
Chris Liu, Liang Zeng, Jiacai Liu, Rui Yan, Jujie He, Chaojie Wang, Shuicheng Yan, Yang Liu, and Yahui Zhou.
\newblock Skywork-reward: Bag of tricks for reward modeling in llms.
\newblock \emph{ArXiv}, abs/2410.18451, 2024.

\bibitem[Liu et~al.(2025)Liu, Zeng, Xiao, He, Liu, Wang, Yan, Shen, Zhang, Xu, Liu, and Zhou]{Liu2025SkyworkRewardV2SP}
Chris Liu, Liang Zeng, Yuzhen Xiao, Jujie He, Jiacai Liu, Chaojie Wang, Rui Yan, Wei Shen, Fuxiang Zhang, Jiacheng Xu, Yang Liu, and Yahui Zhou.
\newblock Skywork-reward-v2: Scaling preference data curation via human-ai synergy.
\newblock 2025.

\bibitem[Lu et~al.(2024)Lu, Yuan, Lin, Lin, Yuan, Zhou, and Zhou]{lu-etal-2024-routing}
Keming Lu, Hongyi Yuan, Runji Lin, Junyang Lin, Zheng Yuan, Chang Zhou, and Jingren Zhou.
\newblock Routing to the expert: Efficient reward-guided ensemble of large language models.
\newblock In Kevin Duh, Helena Gomez, and Steven Bethard (eds.), \emph{Proceedings of the 2024 Conference of the North American Chapter of the Association for Computational Linguistics: Human Language Technologies (Volume 1: Long Papers)}, pp.\  1964--1974, Mexico City, Mexico, June 2024. Association for Computational Linguistics.

\bibitem[MAA(2024)]{AIME2024}
MAA.
\newblock American invitational mathematics examination.
\newblock \emph{https://maa.org/math-competitions/ american-invitational-mathematics-examination-aime.}, 2024.

\bibitem[Madaan et~al.(2023)Madaan, Tandon, Gupta, Hallinan, Gao, Wiegreffe, Alon, Dziri, Prabhumoye, Yang, Gupta, Majumder, Hermann, Welleck, Yazdanbakhsh, and Clark]{madaan2023selfrefine}
Aman Madaan, Niket Tandon, Prakhar Gupta, Skyler Hallinan, Luyu Gao, Sarah Wiegreffe, Uri Alon, Nouha Dziri, Shrimai Prabhumoye, Yiming Yang, Shashank Gupta, Bodhisattwa~Prasad Majumder, Katherine Hermann, Sean Welleck, Amir Yazdanbakhsh, and Peter Clark.
\newblock Self-refine: Iterative refinement with self-feedback.
\newblock In \emph{Thirty-seventh Conference on Neural Information Processing Systems}, 2023.

\bibitem[Mark~Chen(2021)]{humaneval}
et~al. Mark~Chen.
\newblock Evaluating large language models trained on code.
\newblock \emph{ArXiv}, abs/2107.03374, 2021.

\bibitem[OpenAI(2025)]{gpt-4.1}
OpenAI.
\newblock Introducing gpt-4.1 in the api.
\newblock \emph{Accessed: 2025-05-07}, 2025.

\bibitem[Peng et~al.(2023)Peng, Quesnelle, Fan, and Shippole]{peng2023yarn}
Bowen Peng, Jeffrey Quesnelle, Honglu Fan, and Enrico Shippole.
\newblock Yarn: Efficient context window extension of large language models.
\newblock \emph{arXiv preprint arXiv:2309.00071}, 2023.

\bibitem[Rein et~al.(2024)Rein, Hou, Stickland, Petty, Pang, Dirani, Michael, and Bowman]{rein2024gpqa}
David Rein, Betty~Li Hou, Asa~Cooper Stickland, Jackson Petty, Richard~Yuanzhe Pang, Julien Dirani, Julian Michael, and Samuel~R Bowman.
\newblock Gpqa: A graduate-level google-proof q\&a benchmark.
\newblock In \emph{First Conference on Language Modeling}, 2024.

\bibitem[Shnitzer et~al.(2024)Shnitzer, Ou, Silva, Soule, Sun, Solomon, Thompson, and Yurochkin]{shnitzer2024large}
Tal Shnitzer, Anthony Ou, M{\'\i}rian Silva, Kate Soule, Yuekai Sun, Justin Solomon, Neil Thompson, and Mikhail Yurochkin.
\newblock Large language model routing with benchmark datasets, 2024.

\bibitem[Snell et~al.(2025)Snell, Lee, Xu, and Kumar]{bestofn}
Charlie~Victor Snell, Jaehoon Lee, Kelvin Xu, and Aviral Kumar.
\newblock Scaling {LLM} test-time compute optimally can be more effective than scaling parameters for reasoning.
\newblock In \emph{The Thirteenth International Conference on Learning Representations}, 2025.

\bibitem[Srivatsa et~al.(2024)Srivatsa, Maurya, and Kochmar]{srivatsa-etal-2024-harnessing}
Kv~Aditya Srivatsa, Kaushal Maurya, and Ekaterina Kochmar.
\newblock Harnessing the power of multiple minds: Lessons learned from {LLM} routing.
\newblock In Shabnam Tafreshi, Arjun Akula, Jo{\~a}o Sedoc, Aleksandr Drozd, Anna Rogers, and Anna Rumshisky (eds.), \emph{Proceedings of the Fifth Workshop on Insights from Negative Results in NLP}, pp.\  124--134, Mexico City, Mexico, June 2024. Association for Computational Linguistics.

\bibitem[Team et~al.(2024)Team, Mesnard, Hardin, Dadashi, Bhupatiraju, Pathak, Sifre, Rivi{\`e}re, Kale, Love, et~al.]{team2024gemma}
Gemma Team, Thomas Mesnard, Cassidy Hardin, Robert Dadashi, Surya Bhupatiraju, Shreya Pathak, Laurent Sifre, Morgane Rivi{\`e}re, Mihir~Sanjay Kale, Juliette Love, et~al.
\newblock Gemma: Open models based on gemini research and technology.
\newblock \emph{arXiv preprint arXiv:2403.08295}, 2024.

\bibitem[Team(2024{\natexlab{a}})]{qwen2.5}
Qwen Team.
\newblock Qwen2.5: A party of foundation models, September 2024{\natexlab{a}}.

\bibitem[Team(2024{\natexlab{b}})]{qwq-32b-preview}
Qwen Team.
\newblock Qwq: Reflect deeply on the boundaries of the unknown, November 2024{\natexlab{b}}.

\bibitem[Team(2025)]{qwq32b}
Qwen Team.
\newblock Qwq-32b: Embracing the power of reinforcement learning, March 2025.

\bibitem[Touvron et~al.(2023)Touvron, Lavril, Izacard, Martinet, Lachaux, Lacroix, Rozi{\`e}re, Goyal, Hambro, Azhar, Rodriguez, Joulin, Grave, and Lample]{Touvron2023LLaMAOA}
Hugo Touvron, Thibaut Lavril, Gautier Izacard, Xavier Martinet, Marie-Anne Lachaux, Timoth{\'e}e Lacroix, Baptiste Rozi{\`e}re, Naman Goyal, Eric Hambro, Faisal Azhar, Aur'elien Rodriguez, Armand Joulin, Edouard Grave, and Guillaume Lample.
\newblock Llama: Open and efficient foundation language models.
\newblock \emph{ArXiv}, abs/2302.13971, 2023.

\bibitem[Wang et~al.(2025)Wang, WANG, Athiwaratkun, Zhang, and Zou]{wang2025mixtureofagents}
Junlin Wang, Jue WANG, Ben Athiwaratkun, Ce~Zhang, and James Zou.
\newblock Mixture-of-agents enhances large language model capabilities.
\newblock In \emph{The Thirteenth International Conference on Learning Representations}, 2025.

\bibitem[Wang et~al.(2023)Wang, Wei, Schuurmans, Le, Chi, Narang, Chowdhery, and Zhou]{wang2023selfconsistency}
Xuezhi Wang, Jason Wei, Dale Schuurmans, Quoc~V Le, Ed~H. Chi, Sharan Narang, Aakanksha Chowdhery, and Denny Zhou.
\newblock Self-consistency improves chain of thought reasoning in language models.
\newblock In \emph{The Eleventh International Conference on Learning Representations}, 2023.

\bibitem[Wang et~al.(2024{\natexlab{a}})Wang, Bukharin, Delalleau, Egert, Shen, Zeng, Kuchaiev, and Dong]{inform}
Zhilin Wang, Alexander Bukharin, Olivier Delalleau, Daniel Egert, Gerald Shen, Jiaqi Zeng, Oleksii Kuchaiev, and Yi~Dong.
\newblock Helpsteer2-preference: Complementing ratings with preferences.
\newblock \emph{ArXiv}, abs/2410.01257, 2024{\natexlab{a}}.

\bibitem[Wang et~al.(2024{\natexlab{b}})Wang, Liu, Liu, Yao, Huang, He, Li, Li, Che, Zhang, Wang, Wang, Pu, Xu, Fang, Zhao, Zhang, Huang, Lu, Peng, Zheng, Wang, Yang, he, Jiang, Xie, Zhang, Li, Shi, Fu, Zhang, Huang, Xiong, Zhang, Wang, and Song]{wang2024telechat}
Zihan Wang, Xinzhang Liu, Shixuan Liu, Yitong Yao, Yuyao Huang, Zhongjiang He, Xuelong Li, Yongxiang Li, Zhonghao Che, Zhaoxi Zhang, Yan Wang, Xin Wang, Luwen Pu, Huihan Xu, Ruiyu Fang, Yu~Zhao, Jie Zhang, Xiaomeng Huang, Zhilong Lu, Jiaxin Peng, Wenjun Zheng, Shiquan Wang, Bingkai Yang, Xuewei he, Zhuoru Jiang, Qiyi Xie, Yanhan Zhang, Zhongqiu Li, Lingling Shi, Weiwei Fu, Yin Zhang, Zilu Huang, Sishi Xiong, Yuxiang Zhang, Chao Wang, and Shuangyong Song.
\newblock Telechat technical report, 2024{\natexlab{b}}.

\bibitem[Wei et~al.(2022)Wei, Wang, Schuurmans, Bosma, brian ichter, Xia, Chi, Le, and Zhou]{wei2022chain}
Jason Wei, Xuezhi Wang, Dale Schuurmans, Maarten Bosma, brian ichter, Fei Xia, Ed~H. Chi, Quoc~V Le, and Denny Zhou.
\newblock Chain of thought prompting elicits reasoning in large language models.
\newblock In Alice~H. Oh, Alekh Agarwal, Danielle Belgrave, and Kyunghyun Cho (eds.), \emph{Advances in Neural Information Processing Systems}, 2022.

\bibitem[Yang et~al.(2024{\natexlab{a}})Yang, Zhang, Hui, Gao, Yu, Li, Liu, Tu, Zhou, Lin, Lu, Xue, Lin, Liu, Ren, and Zhang]{qwen72brm}
An~Yang, Beichen Zhang, Binyuan Hui, Bofei Gao, Bowen Yu, Chengpeng Li, Dayiheng Liu, Jianhong Tu, Jingren Zhou, Junyang Lin, Keming Lu, Mingfeng Xue, Runji Lin, Tianyu Liu, Xingzhang Ren, and Zhenru Zhang.
\newblock Qwen2.5-math technical report: Toward mathematical expert model via self-improvement.
\newblock \emph{ArXiv}, abs/2409.12122, 2024{\natexlab{a}}.

\bibitem[Yang et~al.(2024{\natexlab{b}})Yang, Yang, Zhang, Hui, Zheng, Yu, Li, Liu, Huang, Dong, Wei, Lin, Yang, Tu, Zhang, Yang, Yang, Zhou, Lin, Dang, Lu, Bao, Yang, Yu, Li, Xue, Zhang, Zhu, Men, Lin, Li, Xia, Ren, Ren, Fan, Su, Zhang, Wan, Liu, Cui, Zhang, Qiu, Quan, and Wang]{Yang2024Qwen25TR}
Qwen~An Yang, Baosong Yang, Beichen Zhang, Binyuan Hui, Bo~Zheng, Bowen Yu, Chengyuan Li, Dayiheng Liu, Fei Huang, Guanting Dong, Haoran Wei, Huan Lin, Jian Yang, Jianhong Tu, Jianwei Zhang, Jianxin Yang, Jiaxin Yang, Jingren Zhou, Junyang Lin, Kai Dang, Keming Lu, Keqin Bao, Kexin Yang, Le~Yu, Mei Li, Mingfeng Xue, Pei Zhang, Qin Zhu, Rui Men, Runji Lin, Tianhao Li, Tingyu Xia, Xingzhang Ren, Xuancheng Ren, Yang Fan, Yang Su, Yi-Chao Zhang, Yunyang Wan, Yuqi Liu, Zeyu Cui, Zhenru Zhang, Zihan Qiu, Shanghaoran Quan, and Zekun Wang.
\newblock Qwen2.5 technical report.
\newblock \emph{ArXiv}, abs/2412.15115, 2024{\natexlab{b}}.

\bibitem[Yao et~al.(2023)Yao, Yu, Zhao, Shafran, Griffiths, Cao, and Narasimhan]{yao2023tree}
Shunyu Yao, Dian Yu, Jeffrey Zhao, Izhak Shafran, Thomas~L. Griffiths, Yuan Cao, and Karthik~R Narasimhan.
\newblock Tree of thoughts: Deliberate problem solving with large language models.
\newblock In \emph{Thirty-seventh Conference on Neural Information Processing Systems}, 2023.

\bibitem[Zeng et~al.(2025)Zeng, Jiang, Wang, Nie, Chen, and Chen]{Zeng2025ACECODERAC}
Huaye Zeng, Dongfu Jiang, Haozhe Wang, Ping Nie, Xiaotong Chen, and Wenhu Chen.
\newblock Acecoder: Acing coder rl via automated test-case synthesis.
\newblock In \emph{Annual Meeting of the Association for Computational Linguistics}, 2025.

\bibitem[Zhang et~al.(2025)Zhang, Zheng, Wu, Zhang, Lin, Yu, Liu, Zhou, and Lin]{qwen7bprm}
Zhenru Zhang, Chujie Zheng, Yangzhen Wu, Beichen Zhang, Runji Lin, Bowen Yu, Dayiheng Liu, Jingren Zhou, and Junyang Lin.
\newblock The lessons of developing process reward models in mathematical reasoning.
\newblock In \emph{Annual Meeting of the Association for Computational Linguistics}, 2025.

\bibitem[Zhou et~al.(2023)Zhou, Lu, Mishra, Brahma, Basu, Luan, Zhou, and Hou]{ifeval}
Jeffrey Zhou, Tianjian Lu, Swaroop Mishra, Siddhartha Brahma, Sujoy Basu, Yi~Luan, Denny Zhou, and Le~Hou.
\newblock Instruction-following evaluation for large language models.
\newblock \emph{arXiv preprint arXiv:2311.07911}, 2023.

\end{thebibliography}
\bibliographystyle{iclr2026_conference}

\newpage
\appendix
\section*{Appendix}

\renewcommand\thefigure{\Alph{figure}} 
\renewcommand\thetable{\Alph{table}} 

\renewcommand\thealgorithm{\Alph{algorithm}} 

\noindent This supplementary document is organized as follows:
\begin{itemize}
    \item Section~\ref{llm_use} contains details on our use of Large Language Models. 
    \item Section~\ref{dataset} contains more details on our experiment datasets. 
    \item Section~\ref{model} contains more details on LLMs and reward models we use for the experiment.
    \item Section~\ref{imple} contains our implementation details.
    \item Section~\ref{cost} contains experiment results on cost efficiency.
    \item Section~\ref{res} contains more comparison experiment results.
    \item Section~\ref{case} contains results for the analysis on specific cases from our approaches.
    \item Section~\ref{qb} contains experiment results for the analysis on question pool.
    
    \item Section~\ref{algorithm} contains more details on our Agent Collaboration Search.
    \item Section~\ref{prompt} contains details on our prompts for each of the seven benchmarks.
\end{itemize}

\section{Details on the use of f Large Language Models}
\label{llm_use}
In this paper, we employ a large language model (LLM) to assist our writing, primarily for aiding or polishing the paper and no other applications are included. Specifically, we use DeepSeek of chat-version via this website: \url{https://yuanbao.tencent.com/chat}. Our usage of the LLM is limited to the following purposes: (1) translating terms and sentences, and (2) refining the phrasing of the manuscript. Finally, we acknowledge the convenient user-interactive LLM service provided by Tencent Yuan Bao based on DeepSeek.

\section{Details on Dataset}
\label{dataset}
In our experiments, we assess the effectiveness of our proposed CTTS-MM across seven diverse benchmarks covering mathematical reasoning, complex QA, instruction following, and code generation. Note that, except for HumanEval~\cite{humaneval}, all datasets are split into test and validation sets, with the validation sets utilized to construct the question pool.
For MBPP~\cite{austin2021mbpp}, we retain the original test set and merge the training and validation sets to serve as the validation split. Specifically, the validation set consists of 464 samples while the test set contains 500 samples.
For LiveCodeBench~\cite{jain2024livecodebench}, we utilize their v5 version as the test set, reserving v6 for validation. 
For MATH~\cite{math500}, we evaluate on the MATH-500 subset and randomly sample 1,000 samples from the original dataset for validation. For AIME~\cite{AIME2024}, we use the 2024 competition problems as the test set, leveraging historical questions (1983–2023) for validation.
For GPQA~\cite{rein2024gpqa}, we adopt the diamond subset consisting of graduate-level science questions as the test set, with the rest used for validation. 
In the IFEval~\cite{ifeval}, 300 instruction-following samples are selected at random for testing, with 241 used for validation.
Finally, for Human-eval, we simply use their original version for test split (164 samples) and no validation split is constructed as mentioned.

\section{Details on LLMs and RMs}
\label{model}

\subsection{LLM Usage}
As we mentioned in our manuscript, we assemble a set of ten mid-sized open-source
LLMs (ranging from 20B to 72B parameters) from diverse architecture families. Specifically, the selected LLMs include: Qwen2.5-32B-Instruct~\cite{qwen2.5}, Qwen-2.5-72B-Instruct~\cite{qwen2.5}, Qwen2.5-Coder-32B-Instruct~\cite{hui2024qwen2}, GLM-Z1-32B-0414~\cite{glm2024chatglm}, DeepSeek-R1-Distill-Qwen-32B~\cite{deepseekai2025}, DeepSeek-R1-Distill-Llama-70B~\cite{deepseekai2025}, QwQ-32B~\cite{qwq-32b-preview}, InternLM2.5-20B-Chat~\cite{cai2024internlm2}, Llama-3.3-70B-Instruct~\cite{grattafiori2024llama}, Llama-3.3-Nemotron-Super-49B-v1~\cite{bercovich2025llama}. Note that this pool of 10 LLMs primarily acts as multi-agent in our CTTS-MM framework and is utilized for comparative experiments on TTS paradigms. For comparison experiments against other methods, we additionally include five open-source models: Gemma-3-27b-it~\cite{team2024gemma}, TeleChat2-35B-32K~\cite{wang2024telechat}, EXAONE-Deep-32B~\cite{exaone-deep}, Qwen3-32B~\cite{qwq32b}, HuatuoGPT-o1-72B~\cite{chen2024huatuogpto1medicalcomplexreasoning}. Details are listed in Table~\ref{tab:llm_bank}.

\begin{table}[!th]
\centering
\caption{Details on utilized LLMs.}
\begin{adjustbox}{max width=\textwidth}
\begin{tabular}{l|c|c}
\toprule
Name                             & Size & Type              \\ 
\midrule

TeleChat2-35B-32K                & 35B  & Instruction-tuned \\
GLM-Z1-32B-0414                  & 32B  & Deep Thinking     \\

Qwen-2.5-72B-Instruct            & 72B  & Instruction-tuned \\
Llama-3.3-70B-Instruct           & 70B  & Instruction-tuned \\
DeepSeek-R1-Distill-Llama-70B    & 70B  & Deep Thinking     \\
DeepSeek-R1-Distill-Qwen-32B     & 32B  & Deep Thinking     \\
Gemma-3-27b-it                   & 27B  & Instruction-tuned \\

Qwen2.5-Coder-32B-Instruct       & 32B  & Instruction-tuned \\
Qwen3-32B                        & 32B  & Deep Thinking     \\
Llama-3.3-Nemotron-Super-49B-v1 & 49B  & Deep Thinking     \\
Qwen2.5-32B-Instruct             & 32B  & Instruction-tuned \\
QwQ-32B                          & 32B  & Deep Thinking     \\
EXAONE-Deep-32B                  & 32B  & Deep Thinking     \\

HuatuoGPT-o1-72B                 & 72B  & Deep Thinking     \\
InternLM2.5-20B-Chat             & 20B  & Instruction-tuned \\ \midrule
\end{tabular}
\end{adjustbox}

\label{tab:llm_bank}
\end{table}

\subsection{RM Usage}

We collect eight off-the-shelf reward models for all our experiments. Specifically, the collected reward models include: Qwen2.5-Math-RM-72B~\cite{qwen72brm}, Qwen2.5-Math-PRM-7B~\cite{qwen7bprm}, Skywork-Reward-Gemma-2-27B~\cite{Liu2024SkyworkReward}, INF-ORM-Llama3.1-70B~\cite{inform}, LDL-Reward-Gemma-2-27B-v0.1~\cite{chen2025LDLRewardGemma}, AceCodeRM-32B~\cite{Zeng2025ACECODERAC}, QRM-Gemma-2-27B~\cite{qrmgemma}, Skywork-Reward-V2-Llama-3.1-8B-40M~\cite{Liu2025SkyworkRewardV2SP}. Details are listed in Table~\ref{tab:rm_bank}

\begin{table}[!th]
\centering
\caption{Details on utilized Reward Models.}
\begin{adjustbox}{max width=1.0\columnwidth}
\begin{tabular}{l|c|l|c}
\toprule
Name                             & Size & Base Model & type               \\ 
\midrule

Qwen2.5-Math-RM-72B & 72B & Qwen2.5-Math-72B & ORM  \\
Qwen2.5-Math-PRM-7B & 7B & Qwen2.5-Math-7B-Instruct & PRM  \\
Skywork-Reward-Gemma-2-27B & 27B & Gemma-2-27B-it & ORM  \\
INF-ORM-Llama3.1-70B & 70B & Llama-3.1-70B-Instruct & ORM \\
LDL-Reward-Gemma-2-27B-v0.1 & 27B & Gemma-2-27B-it & ORM \\
AceCodeRM-32B & 32B & Qwen2.5-Coder-32B-Instruct & ORM \\
QRM-Gemma-2-27B & 32B & Gemma-2-27B-it & ORM \\
Skywork-Reward-V2-Llama-3.1-8B-40M & 8B & Llama-3.1-8B-Instruct & ORM \\

\midrule
\end{tabular}
\end{adjustbox}

\label{tab:rm_bank}
\end{table}

\begin{table*}[!th]

    \centering
        \caption{Comparison results of different TTS paradigms on MBPP, MATH-500 and GPQA. MR$^*$ means utilizing our proposed MoR to select reward models. Multi-agent$^*$ means utilizing ten chosen LLMs.}
\begin{adjustbox}{max width=\textwidth}
    \begin{tabular}{lllcccccccc}
        \toprule
        
         \textbf{Setting} &  \textbf{Model} & \textbf{Reward Model} & \textbf{Weight Method} & \textbf{MBPP} & \textbf{MATH-500} & \textbf{GPQA} & \textbf{Avg.} \\ 
         
        \midrule 
        
        & Qwen2.5-32B-Instruct & - & - & 76.00 & 75.60 & 40.91 & 64.17 \\
        
        & Qwen2.5-72b-Instruct & - & - & 75.80 & 78.80 & 45.45 & 66.68 \\
        
        & Llama-3.3-Nemotron-Super-49B-v1 & - & - & 65.40 & 75.20 & 48.48 & 63.03 \\
        
        & Llama-3.3-70B-Instruct & - & - & 70.40 & 73.00 & 46.97 & 63.46 \\
        \multicolumn{1}{l}{\multirow{-5}{2cm}{Single Agent}}
        & DeepSeek-R1-Distill-Llama-70B & - & - & 76.40 & 82.8 & 60.10 & 73.10 \\
        \midrule

        & Qwen2.5-32B-Instruct & AceCodeRM-32B & - & 77.40 & 78.2 & 47.47 & 67.69 \\
          & Qwen2.5-32B-Instruct & Qwen2.5-Math-RM-72B & - & 77.00 & 78.80 & 46.46 & 67.42 \\
        & Qwen2.5-72b-Instruct & AceCodeRM-32B & - & 76.60 & 80.20 & 51.01 & 69.27 \\
        & Qwen2.5-72b-Instruct & Qwen2.5-Math-RM-72B & - & 76.00 & 80.80 & 50.51 & 69.10 \\
        
        &Llama-3.3-Nemotron-Super-49B-v1 & AceCodeRM-32B & - & 66.40 & 76.00 & 50.80 & 64.40 \\
        
        & Llama-3.3-Nemotron-Super-49B-v1 & Qwen2.5-Math-RM-72B & - & 65.80 & 76.80 & 50.00 & 64.20 \\

        & Llama-3.3-70B-Instruct & AceCodeRM-32B & - & 71.20 & 73.40 & 48.00 & 64.20 \\
        & Llama-3.3-70B-Instruct & Qwen2.5-Math-RM-72B & - & 70.80 & 73.80 & 47.47 & 64.02 \\
        
        & DeepSeek-R1-Distill-Llama-70B & AceCodeRM-32B & - & 77.00 & 82.60 & 59.09 & 72.90 \\
        \multicolumn{1}{l}{\multirow{-10}{2cm}{SA-SR}}
        & DeepSeek-R1-Distill-Llama-70B & Qwen2.5-Math-RM-72B & - & 76.60 & 83.20 & 59.09 & 72.96 \\

        \midrule   
        
        & Qwen2.5-32B-Instruct & AceCodeRM-32B+Qwen2.5-Math-RM-72B & softmax & 76.60 & 78.20 & 48.48 & 67.76 \\
        & Qwen2.5-32B-Instruct & AceCodeRM-32B+Qwen2.5-Math-RM-72B & linear & 76.60 & 78.00 & 47.80 & 67.47 \\
        & Qwen2.5-32b-Instruct & MR$^*$ & - & 78.00 & 79.4 & 51.01 & 69.47 \\
        
        & Qwen2.5-72B-Instruct & AceCodeRM-32B+Qwen2.5-Math-RM-72B & softmax & 76.80 & 80.20 & 51.51 & 69.50 \\
        
        & Qwen2.5-72B-Instruct & AceCodeRM-32B+Qwen2.5-Math-RM-72B & linear & 77.00 & 80.20 & 52.02 & 69.74 \\
        
        & Qwen2.5-72b-Instruct & MR$^*$ & - & 77.20 & 81.4 & 53.53 & 70.71 \\

        &Llama-3.3-Nemotron-Super-49B-v1 & AceCodeRM-32B+Qwen2.5-Math-RM-72B & softmax & 66.20 & 76.60 & 51.52 & 64.77 \\
        
        &Llama-3.3-Nemotron-Super-49B-v1 & AceCodeRM-32B+Qwen2.5-Math-RM-72B & linear & 66.20 & 76.40 & 51.52 & 64.70 \\
        
        & Llama-3.3-Nemotron-Super-49B-v1 & MR$^*$ & - & 66.80 & 76.80 & 54.55 & 66.05 \\

        & Llama-3.3-70B-Instruct & AceCodeRM-32B+Qwen2.5-Math-RM-72B & softmax & 71.40 & 74.00 & 48.48 & 64.63 \\
        
        & Llama-3.3-70B-Instruct & AceCodeRM-32B+Qwen2.5-Math-RM-72B & linear & 71.40 & 74.00 & 48.99 & 64.80 \\
        
        & Llama-3.3-70B-Instruct & MR$^*$ & - & 72.00 & 74.40 & 49.49 & 65.30 \\
        
        & DeepSeek-R1-Distill-Llama-70B & AceCodeRM-32B+Qwen2.5-Math-RM-72B & softmax & 76.60 & 83.00 & 60.10 & 73.23 \\
        & DeepSeek-R1-Distill-Llama-70B & AceCodeRM-32B+Qwen2.5-Math-RM-72B & linear & 76.80 & 83.20 & 60.10 & 73.36 \\
        \multicolumn{1}{l}{\multirow{-15}{2cm}{SA-MR}}
        & DeepSeek-R1-Distill-Llama-70B & MR$^*$ & - & 77.20 & 83.60 & 60.60 & 73.80 \\

        \midrule   

        & Multi-agent$^*$ & Skywork-Reward-V2-Llama-3.1-8B-40M & - & 77.00 & \cellcolor{toptwoRed}91.20 & \cellcolor{toptwoRed}61.11 & 75.97 \\
        
        & Multi-agent$^*$ & Qwen2.5-Math-RM-72B & - & \cellcolor{toptwoRed}80.6 & \cellcolor{toponeRed}91.8 & \cellcolor{toptwoRed}61.11 & \cellcolor{toptwoRed}77.84 \\

        & Multi-agent$^*$ & LDL-Reward-Gemma-2-27B-v0.1 & - & 78.80 & 91.00 & 62.63 & 77.48 \\
        
        \multicolumn{1}{l}{\multirow{-4}{2cm}{MA-SR}}
        
        & Multi-agent$^*$ & AceCodeRM-32B & - & \cellcolor{toponeRed}82.2 & 90.8 & \cellcolor{toponeRed}61.62 & \cellcolor{toponeRed}78.21 \\

        \midrule   

        \multicolumn{1}{l}{MA-MR (Proposed CTTS-MM)}& Multi-agent$^*$ & MR$^*$ & - & \textbf{83.20} & \textbf{93.00} & \textbf{64.14} & \textbf{80.11} \\

        \bottomrule
    \end{tabular}
    \end{adjustbox}

    \label{table:paradigm}
\end{table*}

\begin{table*}[th!]

  \centering
      \caption{Main results of  CTTS-MM compared with the optimal setting of Best of N on five benchmarks.}
\begin{adjustbox}{max width=\textwidth}
    \begin{tabular}{lcccccc}
      \toprule
      \textbf{Model} & \textbf{AIME} & \textbf{MATH-500} & \textbf{MBPP} & \textbf{LiveCodeBench}  & \textbf{Human-eval}  & \textbf{Avg} \\
      \midrule
      
      \rowcolor{gray!100} \multicolumn{7}{c}{\textit{Open-source LLMs}} \\

      Qwen-2.5-72B-Instruct & 16.70 & 78.80 & 75.80 & 26.10  & 78.66  & 55.21 \\
      
      DeepSeek-R1-Distill-Llama-70B & 60.00 & 82.80 & 76.40 & 40.70  & 92.07  & 70.39 \\
      Llama-3.3-Nemotron-Super-49B-v1 & 16.70 & 75.20 & 65.40 & 28.00  & 84.76  & 54.01 \\
      QwQ-32B & 46.70 & 87.80 & \cellcolor{toponeRed}81.80 & 38.60  & 92.07   & 69.39 \\
      
      InternLM2.5-20B-Chat & 3.30 & 55.20 & 55.00 & 14.90  & 69.51  & 39.58 \\
      
      Gemma-3-27b-it & 30.00 & 84.00 & 70.40 & 27.70  & 86.59   & 59.74 \\
      
      Qwen2.5-32b-Instruct & 20.00 & 75.60 & 76.00 & 24.00  & 77.44   & 54.61 \\
      
      TeleChat2-35B-32K & 10.00 & 70.00 & 70.00 & 19.50  & 73.17   & 48.53 \\
        EXAONE-Deep-32B & 33.30 & 84.38 & 72.80 & 31.60  & 93.90   & 63.20 \\
     
      GLM-Z1-32B-0414 & \cellcolor{toponeRed}66.70 & \cellcolor{toponeRed}90.00 & 74.40 & 44.40  & \cellcolor{toponeRed}96.34  & \cellcolor{toponeRed}74.37 \\
      Llama-3.3-70B-Instruct & 30.00 & 73.00 & 70.40 & 30.10  & 84.15 & 57.53 \\
      
 Qwen3-32B & 53.30 & 88.00 & 50.60 & 33.40  & 90.85  & 63.23 \\
      Qwen2.5-Coder-32B-Instruct & 16.70 & 73.60 & 78.00 & 27.70  & 87.80  & 56.76 \\
      
      HuatuoGPT-o1-72B & 16.70 & 73.00 & 78.00 & 27.40  & 85.37  & 56.09 \\

      DeepSeek-R1-Distill-Qwen-32B & 56.70 & 85.60 & 81.00 & \cellcolor{toponeRed}44.70  & 95.73  & 72.75 \\

      \rowcolor{gray!100} \multicolumn{7}{c}{\textit{Setting for Best of N}} \\

        Baseline &  66.70  &  90.8   &   75.00  &   44.99   &  96.34     &  74.77 \\
        Optimal Setting &  66.70  &  90.8   &   82.20  &   46.20   &  96.34     &  76.45 \\
    
      \rowcolor{gray!100} \multicolumn{7}{c}{\textit{Ours v.s. Optimal Setting for Best of N}} \\

      \textbf{CTTS-MM(ours)} & \textbf{70.00} & \textbf{93.00} & \textbf{83.20} & \textbf{52.28}  
       & \textbf{97.56} &   \textbf{79.21} \\

      \textit{- v.s. Best of N Baseline} & \positive{3.30} & \positive{2.20} & \positive{8.20} & \positive{7.29}  & \positive{1.22} & \positive{4.82} \\

      \textit{- v.s. Best of N Optimal Setting} & \positive{3.30} & \positive{2.20} & \positive{1.00} & \positive{6.08}  & \positive{1.22} & \positive{2.76} \\
      
            \bottomrule
    \end{tabular}
  \end{adjustbox}

  \label{tab:main_results} 
\end{table*}

\begin{table*}[!th]

\centering
\caption{PRR accuracy of different reward models on four validation datasets.}
\begin{adjustbox}{max width=\textwidth}

\begin{tabular}{lccccc}
\toprule
\textbf{Reward Model} & MBPP-Val  & MATH-Val & AIME-Val & LiveCodeBench-Val  & Avg\\
\midrule
Skywork-Reward-Gemma-2-27B                  & 61.83                                                   & 51.37                            & 50.53                        & 49.19                                                    & 53.23                         \\
LDL-Reward-Gemma-2-27B-v0.1                      & 61.23                                                      & 47.69                            & 43.75                        & 47.13                                                          & 49.95                         \\
Skywork-Reward-V2-Llama-3.1-8B-40M          & 58.44                                                    & 76.75                            & \cellcolor{toptwoRed}82.08                        & 80.72                                                         & 74.50                         \\
INF-ORM-Llama3.1-70B                        & 66.98                                                   & 51.51                            & 47.43                        & 49.29                                                         & 53.80                         \\
Qwen2.5-Math-RM-72B                         & \cellcolor{toptwoRed}68.54                                                  & \cellcolor{toponeRed}87.73                            & \cellcolor{toponeRed}89.13                        & \cellcolor{toptwoRed}86.39                                                          & \cellcolor{toponeRed}82.95 \\
Qwen2.5-Math-PRM-7B & 67.05                                                  & 67.15                            & 49.32                        & 35.74                                                         & 54.82                         \\
AceCodeRM-32B   & \cellcolor{toponeRed}75.00                                                & \cellcolor{toptwoRed}78.73                            & 75.83                        & \cellcolor{toponeRed}88.48                                                        & \cellcolor{toptwoRed}79.51 \\
QRM-Gemma-2-27B                                 & 61.98                                                    & 49.64                            & 45.46                        & 53.08                                                        & 52.54  \\
\bottomrule
\end{tabular}
\end{adjustbox}

\label{rm_acc}
\end{table*}

\section{Implementation Details}
\label{imple}

\subsection{Inference Details}
All experiments are conducted under the same inference settings. We employ VLLM~\cite{kwon2023efficient} as the backend for executing LLM inference. The sampling temperature is fixed at 0.7, and the output sequence is set to 8,192 tokens to prevent excessively long generations. A presence penalty of 1.05 is applied to discourage repetitive outputs. In cases where the input context exceeds the model’s token limit, we apply the YaRN method~\cite{peng2023yarn} to extend the context window. For aggregator, we use Llama-3.3-70B-Instruct. For embedding computation, we adopt Linq-Embed-Mistral~\cite{kim2024linq} across all experiments, with a fixed embedding dimension of 4,096. For reward models, VLLM is also utilized as inference backend except for Qwen2.5-Math-PRM-7B and Skywork-Reward-V2-Llama-3.1-8B-40M (These two reward models are already fast enough using their huggingface version with Flash Attention). As for other RMs, official VLLM only support Qwen2.5-Math-RM-72B. For other RMs used, we implement their VLLM version by ourself (This will be released along with our code). All reward models are set to bfloat16 while other configurations stick to their original settings.

\subsection{Hyperparameters}


For all experiments, we use the same hyperparameters to ensure fair comparison. Specifically, for greedy search process of our ACS, we set top $k=2$ to initialize our search subset while the number of aggregating is set to 8. For expanding the reward model pool, we consider combinations involving 2 and 3 reward models under three weight method: softmax, linear and sum. The selection number k is set to 100 while the tolerance threshold coefficient $\gamma = 0.95$.

\subsection{Details on Related Methods} Besides comparing the performance of single LLMs, we also compare our CTTS-MM with five popular multi-LLMs collaboration methods, and the experimental settings are as follows: Symbolic-MoE~\cite{Chen2025SymbolicMA} retains its original model profiling and LLM selection framework while employing Llama-3.3-70B-Instruct for final response aggregation.
MoA~\cite{wang2025mixtureofagents} employs 15 LLMs as references, also utilizing Llama-3.3-70B-Instruct as the aggregator. For Self Consistency~\cite{chen2024universal}, we select the best LLM on the validation datasets of each benchmark to generate eight responses per query, respectively. Majority Voting~\cite{chen2024are} determines the final output through voting among 15 reference LLMs. For Best of N$^*$~\cite{bestofn}, N is set to 8. And we use GLM-Z1-32B-0414 as our base model which obtains the highest average accuracy in open-source LLMs while Qwen2.5-Math-RM-72B is utilized as reward model for choosing the best answer as it achieves the best average PRR accuracy as shown in Table~\ref{rm_acc}. Like Self Consistency, we also conduct experiments of stronger settings for Best of N, which is shown in Section~\ref{res}.

\subsection{Details on the experiments of inference time}
In the manuscript, we investigate the inference time of our CTTS-MM in comparison to other related methods. Here, we provide more details on the experiment setting. Specifically, we report the average inference time per response on MATH-500 and LiveCodeBench with 10 reference LLMs, each deployed on a server with eight A800 GPUs. Although compared with a single LLM, multi-agent systems basically require more LLM forward passes and need more computational resources, most of these forward passes, e.g., the inferences of different referencers and aggregating multiple times, are independent and can be parallelized, making the
overall inference time primarily determined by the slowest LLM. For a fair comparison, we apply parallel inference acceleration for all related methods.

\begin{figure*}[!ht]
  \centering
  \includegraphics[width=\textwidth]{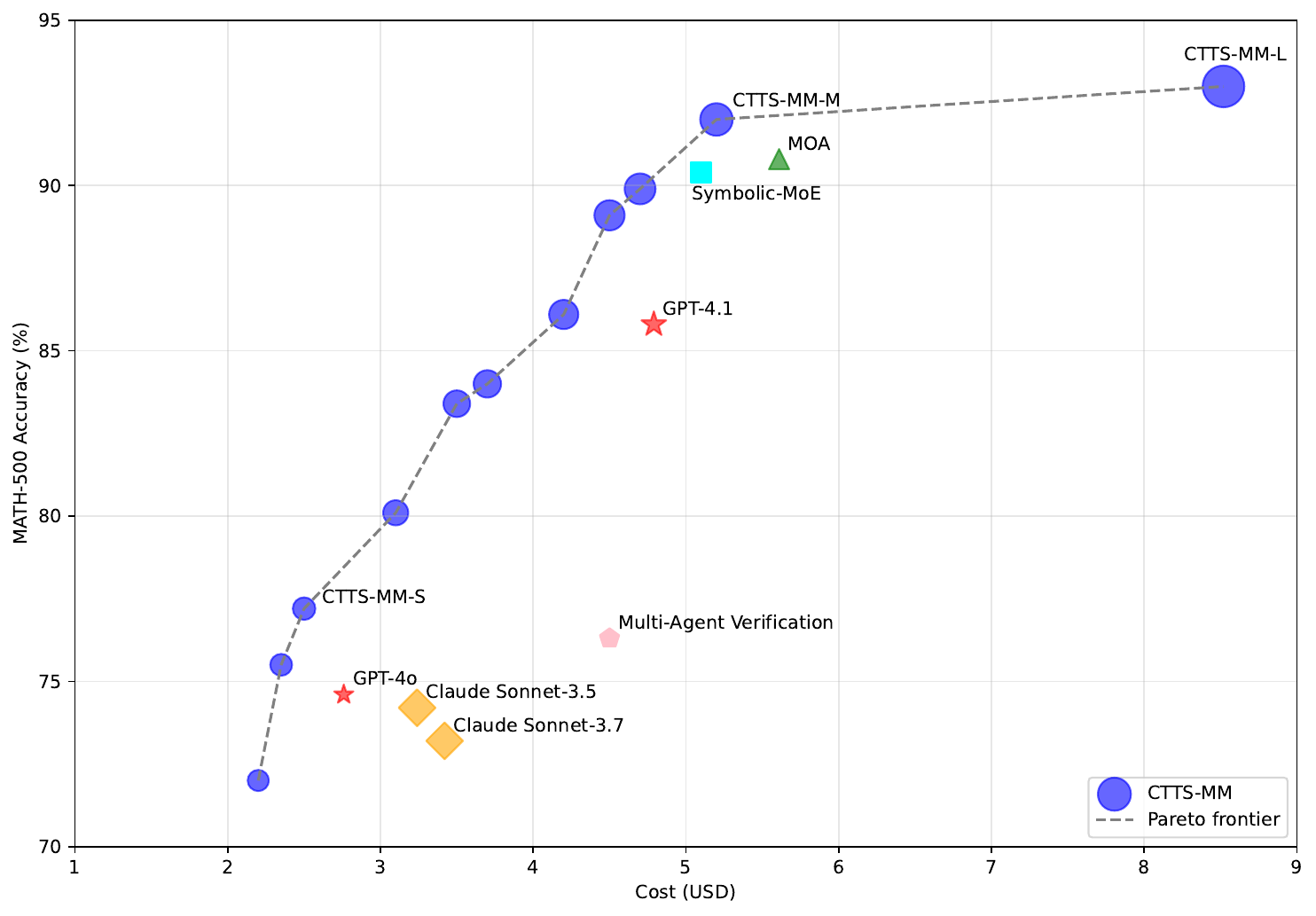}
  \caption{Results of performance versus cost on MATH-500. }
  \label{fig:cost}
\end{figure*}
\section{Results on cost efficiency}
\label{cost}
In the manuscript, we discuss about our analysis on the cost efficiency of out method. Here, we provide qualitative results. As shown in Figure~\ref{fig:cost}, we observe that different variants of our CTTS-MM lie along this frontier, in contrast to proprietary LLMs such as GPT-4.1 and Claude Sonnet-3.5, as well as other multi-agent approaches like MOA~\cite{wang2025mixtureofagents} and Symbolic-MoE~\cite{Chen2025SymbolicMA}, which are not cost-optimal and incur higher expenses for comparable performance. When quality is the primary objective, CTTS-MM-L represents the best configuration. For scenarios requiring a favorable trade-off between quality and cost, CTTS-MM-M achieves competitive cost with multi-agent based methods like MOA and Symbolic-MoE or GPT-4.1 while obtains higher level of quality. Notably, it outperforms GPT-4.1 by approximately 6\% and Symbolic-MoE by around 2\%.  

\section{More Experiment Results }
\label{res}

\subsection{Comparison Results on TTS paradigms}

In our manuscript, we conduct exploratory experiments measuring performance variation among different paradigms. Here, we present additional results on more base models and weight method in Table~\ref{table:paradigm}. Results basically reveal the same conclusion on Llama-3.3-70B-Instruct and DeepSeek-R1-Distill-Llama-70B as CTTS paradigms outperform both STTS and single-
model baselines under most settings.

\subsection{Comparison Results on Five Benchmarks}
We conduct additional experiments to compare our CTTS-MM with Best of N under its optimal setting on AIME, MATH-500, MBPP, LiveCodeBench and Human-eval. The results are shown in Table~\ref{tab:main_results}. For the baseline of Best of N, we keep it the same with our manuscript, where we use GLM-Z1-32B-0414 as our base model which obtains the highest average accuracy in open-source LLMs while Qwen2.5-Math-RM-72B is utilized as reward model for choosing the best answer. As for Optimal Setting, we select the best open-source LLM on the validation datasets of each benchmark while the best reward model is utilized based on results from Table~\ref{rm_acc}. Note that we have no validation split on Human-eval, thus AceCodeRM-32B is utilized since it has best performance on coding benchmark. Results consistently show that our CTTS-MM superior performance increase, still outperforming the optimal setting of Best of N across all five benchmarks and by +2.76\% on average accuracy. 

\section{Case Study}
\label{case}
\begin{figure*}[!ht]
  \centering
  \includegraphics[width=\textwidth]{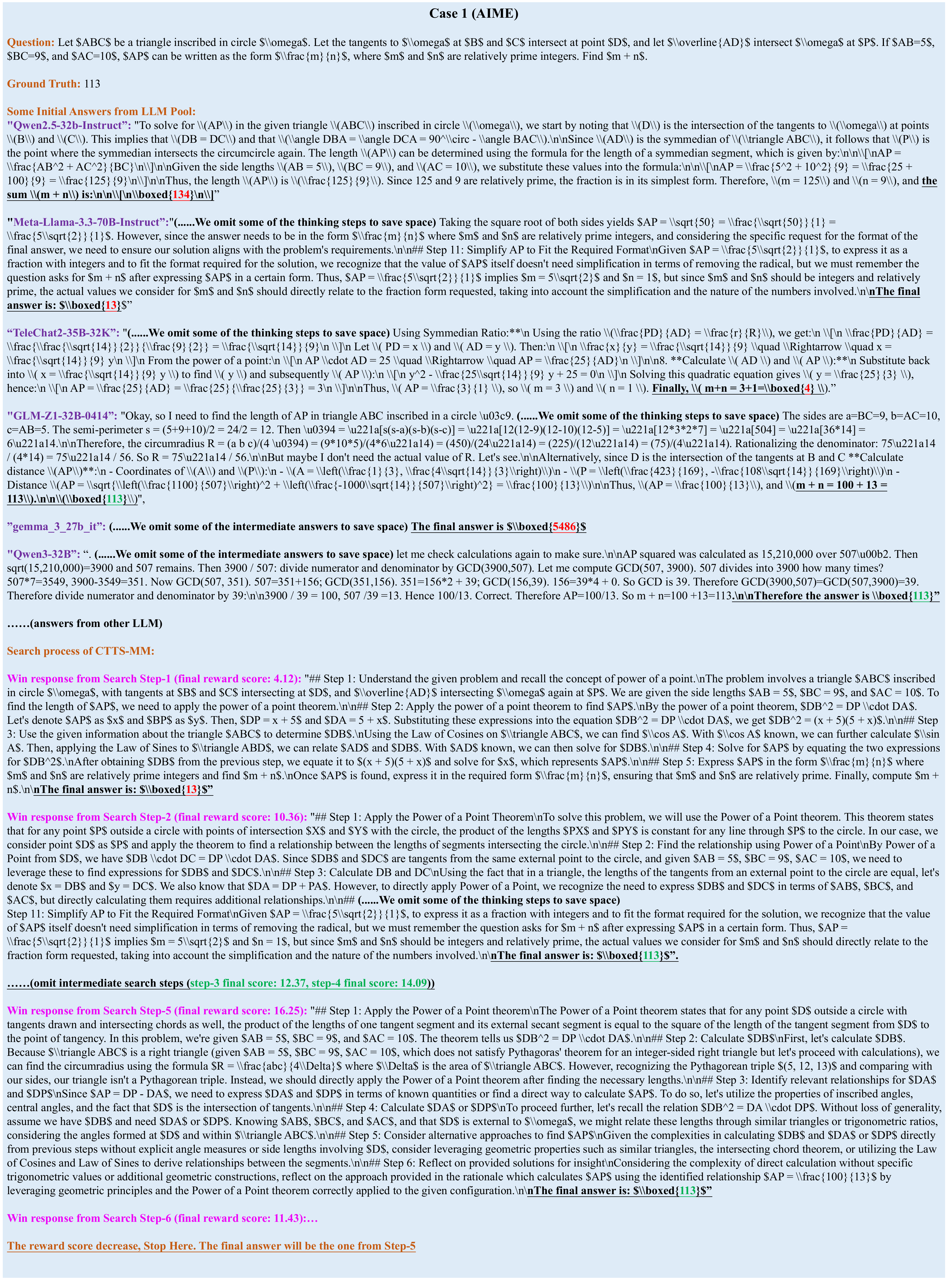}
  \caption{Detailed case results from AIME.}
  \label{fig:case1}
\end{figure*}

\begin{figure*}[!ht]
  \centering
  \includegraphics[width=1.0\textwidth]{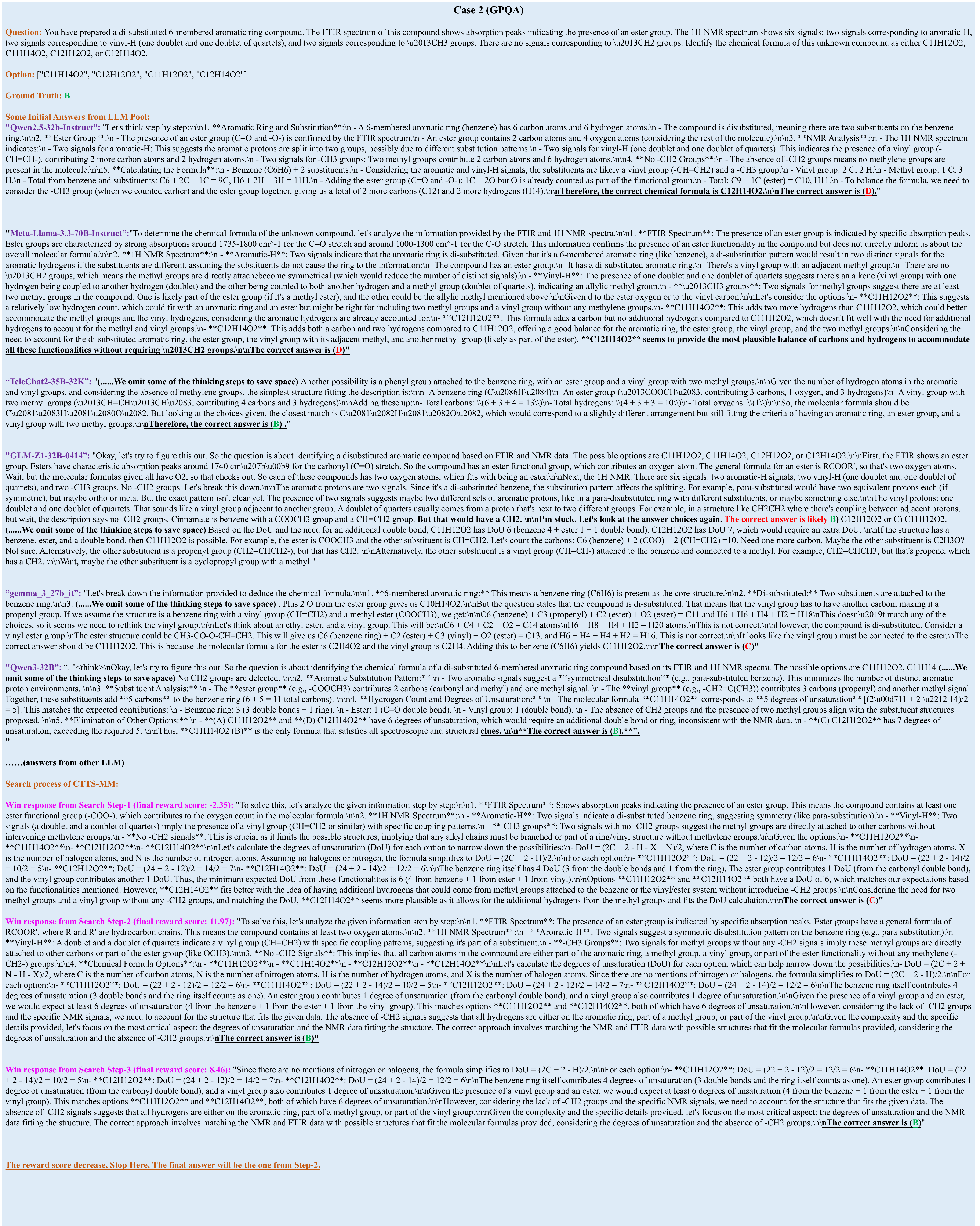}
  \caption{Detailed case results from GPQA.}
  \label{fig:case2}
\end{figure*}

\begin{figure*}[!ht]
  \centering
  \includegraphics[width=\textwidth]{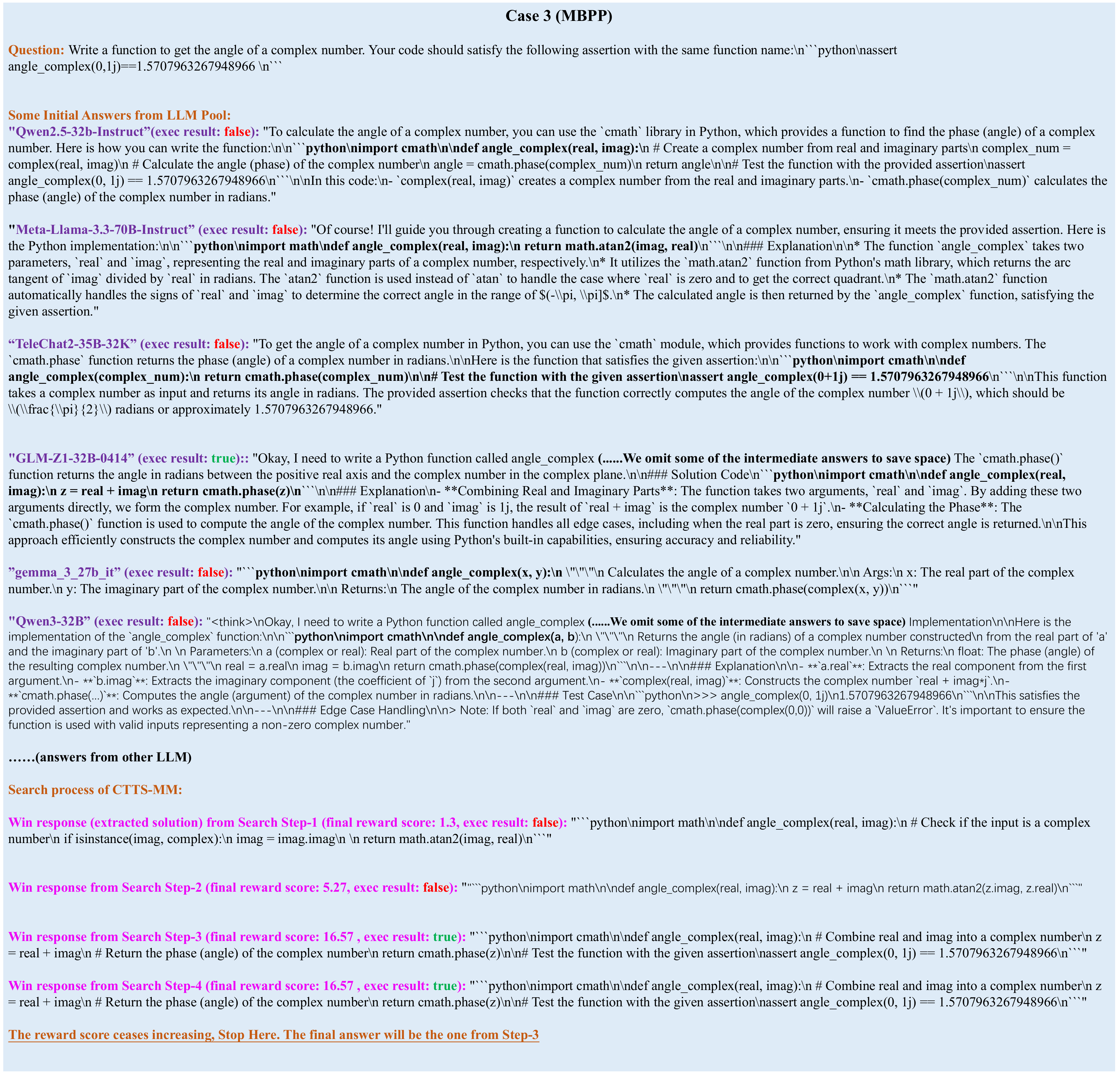}
  \caption{Detailed case results from MBPP.}
  \label{fig:case3}
\end{figure*}

We further study the detailed case from our experiments. The detailed contents are provided in Figure~\ref{fig:case1},~\ref{fig:case2} and~\ref{fig:case3}. We observe that our CTTS-MM is capable of deriving the correct answer through greedy search over the LLM pool using reward scores provided by MoR, even when only one or two models initially produce the correct answer. This result highlights the efficacy of our proposed search process guided by reward scores and underscores the crucial role of our MoR method in providing precise rewards.

\begin{table}[!t]
\centering
\caption{Comparison results of cross-domain question pools.}
\begin{adjustbox}{max width=1.0\textwidth}
\begin{tabular}{lcccc}
\toprule
Question Pool & AIME & MBPP & MATH-500 & LiveCodeBench \\
\midrule
MATH-Val & 66.67 & 80.1 & 92.2 & 48.94 \\
MBPP-Val & 66.67 & 82.8 & 91.8 & 50.15 \\
All (Seven Datasets) & 70.00 & 83.2 & 93.00 & 52.28 \\
\bottomrule
\end{tabular}
\end{adjustbox}

\label{tab:qbb}
\end{table}

\section{Analysis on question pool}
\label{qb}
We further investigate the impact of utilizing question pools out of domains on overall performance. As shown in Table~\ref{tab:qbb}, the performance degradation caused by employing out-of-domain question pools in MoR remains marginal. For instance, using an out-of-domain dataset (MBPP) as the question pool for evaluation on a math-related dataset (MATH) results in a marginal performance decrease by 0.4\% compared to using an in-domain dataset as the question pool. Similar trends can be observed among other datasets, demonstrating the robustness and stability of our MoR approach. Moreover, when comparing against using a combined question pool from all datasets, we observe consistent performance improvements, highlighting the strong scalability of the question pool.

\section{More Details on ACS}
\label{algorithm}
Algorithm~\ref{alg:greedy} shows the detailed algorithm of our Agent Collaboration Search. Our ACS employs a reward-guided greedy search with early stopping and residual aggregation. Given a set of $n$ candidate answers, ACS first ranks them using a Mixture of Reward Models (MoR) and selects the top-$k$ answers to initialize the search set. An aggregator then combines these into a composite answer, whose reward score is evaluated. Iteratively, ACS examines whether adding remaining candidates improves the aggregated answer. The search terminates when no further improvement is found. Finally, a residual aggregation step merges the best answer with all initial candidates to mitigate information loss, and the output with the higher reward is selected.

\begin{algorithm}[ht!]
	\caption{Greedy Search Paradigm of ACS}
	\label{alg:greedy}
	\begin{algorithmic}[1]
        \REQUIRE {Question $q$, LLM set $D_A$, An initial Answer Set $A_0$, Reward Model set $D_R$, Question Pool $Q_p$, MOR Selective Fuction $F$, Search Steps $T$, Aggregator $Agg$, Initial Search Set Num $k$.}
        \ENSURE The optimal answer $A$ to question $q$.
        \FOR{$M$ in $D_A$}
            \STATE $A_0.add(M(q))$ \hfill \COMMENT{\textbf{Initialize the answer set.}}
        \ENDFOR
        \STATE $MOR \leftarrow F(q, Q_p, D_R)$ \hfill \COMMENT{\textbf{Select suitable reward models or their combinations.}}
        \STATE $Score_0 \leftarrow MOR(A_0)$
        \FOR {$i=1$ to $T$}
            \IF {$i==0$}
                \STATE $chosen\_index \leftarrow Score_0.topk(k).index$ 
                \STATE $best\_ans\_set \leftarrow A_0[chosen\_index]$ \hfill \COMMENT{\textbf{Top k answers ranked by their reward scores constitute the initial search subset.}}
                \STATE $current\_best \leftarrow Agg(best\_ans\_set)$ \hfill \COMMENT{\textbf{Aggregate the above top k answers.}}
                \STATE $current\_best\_score \leftarrow MOR(current\_best)$
            \ELSE
                \IF{$(A_0-best\_ans\_set).empty()$}
                    \STATE $break$
                \ENDIF
                \STATE $improvement \leftarrow False$
                \STATE $ans\_to\_be\_searched \leftarrow A_0-best\_ans\_set$ 
                \FOR{$ans$ in $ans\_to\_be\_searched$} 
                    \STATE $temp\_candidate \leftarrow best\_ans\_set$   \hfill \COMMENT{\textbf{Search for answers with improvement.}}
                    \STATE $temp\_candidate.add(ans)$
                    \STATE $temp\_ans \leftarrow Agg(temp\_candidate)$
                    \STATE $temp\_ans\_score \leftarrow MOR(temp\_ans)$
                    \IF {$temp\_ans\_score\succ current\_best\_score$}
                        \STATE $best\_ans\_set \leftarrow temp\_candidate$
                        \STATE $current\_best \leftarrow temp\_ans$
                        \STATE $current\_best\_score \leftarrow temp\_ans\_score$
                        \STATE $improvement \leftarrow True$
                    \ENDIF
                \ENDFOR
                \IF {$!improvement$} 
                    \STATE $break$  \hfill \COMMENT{\textbf{If no improvement, stop.}}
                \ENDIF
                
            \ENDIF
        \ENDFOR
        \IF {$residual$} 
            \STATE $residual\_set \leftarrow \mathcal{A_0} \cup \{current\_best\}$  \hfill \COMMENT{\textbf{Residual Aggregation}}
            \STATE $residual\_answer \leftarrow Agg(residual\_set)$
            \STATE $residual\_score \leftarrow MoR(residual\_answe)$
            \IF {{$residual\_score \succ current\_best\_score$}}
                \STATE $current\_best \leftarrow residual\_answer$
            \ENDIF
        \ENDIF
        \STATE $A \leftarrow current\_best$
            
	\end{algorithmic}
\end{algorithm}

\section{Details on Prompt}
\label{prompt}
\begin{figure*}[t]
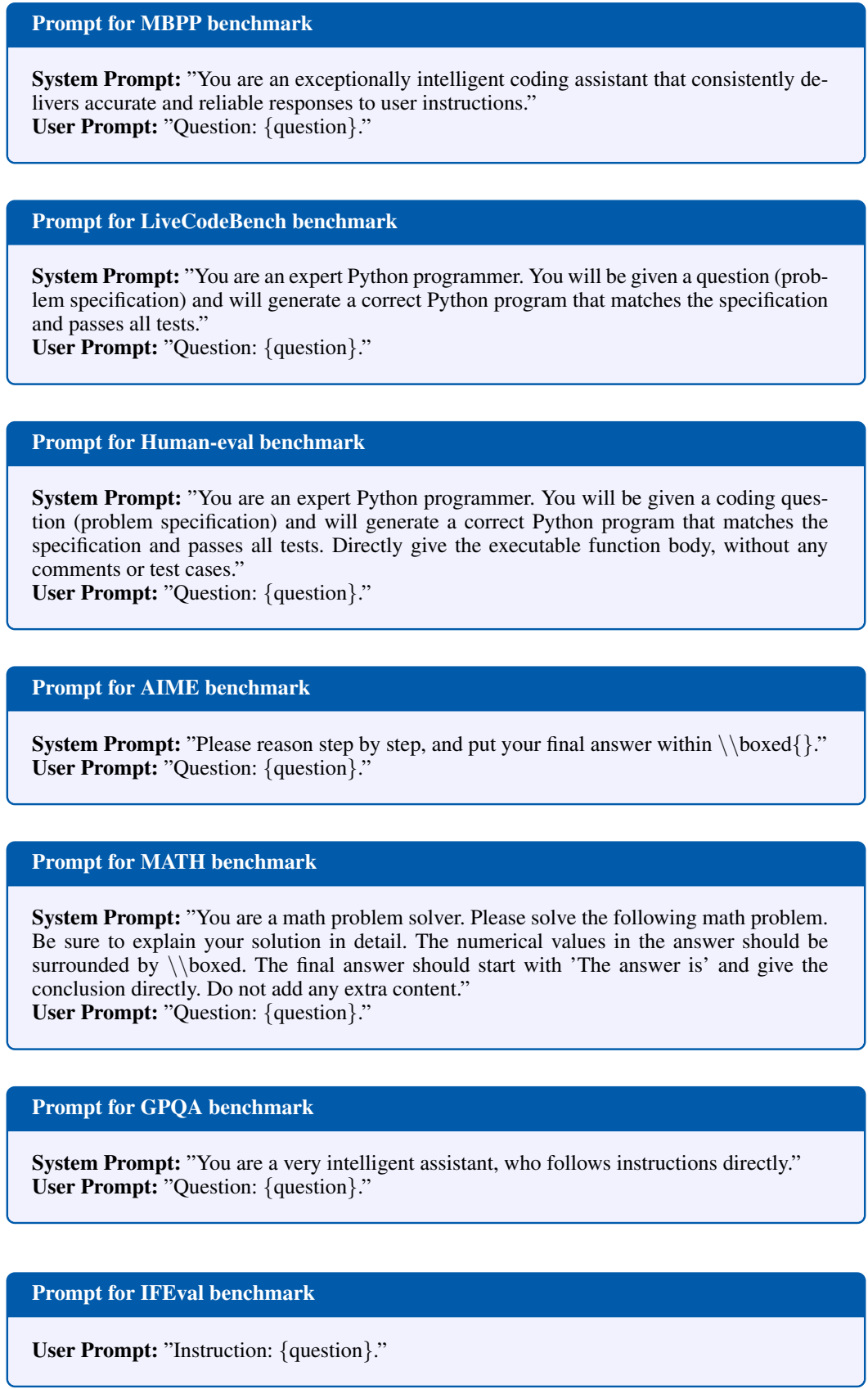
  
\centering

\begin{promptbox3}
\textbf{System Prompt:} "You are an exceptionally intelligent coding assistant that consistently delivers accurate and reliable responses to user instructions." \\
\textbf{User Prompt:} "Question: \{question\}."
\end{promptbox3}
\vspace{2mm}
\begin{promptbox4}
\textbf{System Prompt:} "You are an expert Python programmer. You will be given a question (problem specification) and will generate a correct Python program that matches the specification and passes all tests." \\
\textbf{User Prompt:} "Question: \{question\}."
\end{promptbox4}
\vspace{2mm}
\begin{promptbox6}
\textbf{System Prompt:} "You are an expert Python programmer. You will be given a coding question (problem specification) and will generate a correct Python program that matches the specification and passes all tests. Directly give the executable function body, without any comments or test cases." \\
\textbf{User Prompt:} "Question: \{question\}."
\end{promptbox6}

\vspace{2mm}
\begin{promptbox1}
\textbf{System Prompt:} "Please reason step by step, and put your final answer within \textbackslash\textbackslash boxed\{\}." \\
\textbf{User Prompt:} "Question: \{question\}."
\end{promptbox1}
\vspace{2mm}

\begin{promptbox2}
\textbf{System Prompt:} "You are a math problem solver. Please solve the following math problem. Be sure to explain your solution in detail. The numerical values in the answer should be surrounded by \textbackslash\textbackslash boxed{}. The final answer should start with 'The answer is' and give the conclusion directly. Do not add any extra content." \\
\textbf{User Prompt:} "Question: \{question\}."
\end{promptbox2}
\vspace{2mm}

\begin{promptbox5}
\textbf{System Prompt:} "You are a very intelligent assistant, who follows instructions directly." \\
\textbf{User Prompt:} "Question: \{question\}."
\end{promptbox5}
\vspace{2mm}

\vspace{2mm}
\begin{promptbox7}
\textbf{User Prompt:} "Instruction: \{question\}."
\end{promptbox7}
\vspace{2mm}

\vspace{-2mm}
\caption{Prompts for seven benchmarks.}
\label{prompt_dataset}
\end{figure*}

To obtain the optimal task-specific performance across heterogeneous benchmarks, we design prompt individually for each of the seven benchmarks, taking into account their unique characteristics, as shown in Figure~\ref{prompt_dataset} Moreover, the design of the aggregator prompt within our CTTS-MM framework is refined based on MOA~\cite{wang2025mixtureofagents}, as illustrated in Figure~\ref{prompt_aggregator}.

\newtcolorbox{promptbox1}{
  colback=blue!5!white,
  colframe=cyan!30!blue,
  fonttitle=\bfseries,
  title=Prompt for AIME benchmark,
  boxrule=1pt,
  arc=3pt,
  boxsep=5pt,
  left=6pt,  
  enhanced jigsaw  
}

\newtcolorbox{promptbox2}{
  colback=blue!5!white,
  colframe=cyan!30!blue,
  fonttitle=\bfseries,
  title=Prompt for MATH benchmark,
  boxrule=1pt,
  arc=3pt,
  boxsep=5pt,
  left=6pt,  
  enhanced jigsaw  
}

\newtcolorbox{promptbox3}{
  colback=blue!5!white,
  colframe=cyan!30!blue,
  fonttitle=\bfseries,
  title=Prompt for MBPP benchmark,
  boxrule=1pt,
  arc=3pt,
  boxsep=5pt,
  left=6pt,  
  enhanced jigsaw  
}

\newtcolorbox{promptbox4}{
  colback=blue!5!white,
  colframe=cyan!30!blue,
  fonttitle=\bfseries,
  title=Prompt for LiveCodeBench benchmark,
  boxrule=1pt,
  arc=3pt,
  boxsep=5pt,
  left=6pt,  
  enhanced jigsaw  
}

\newtcolorbox{promptbox5}{
  colback=blue!5!white,
  colframe=cyan!30!blue,
  fonttitle=\bfseries,
  title=Prompt for GPQA benchmark,
  boxrule=1pt,
  arc=3pt,
  boxsep=5pt,
  left=6pt,  
  enhanced jigsaw  
}

\newtcolorbox{promptbox6}{
  colback=blue!5!white,
  colframe=cyan!30!blue,
  fonttitle=\bfseries,
  title=Prompt for Human-eval benchmark,
  boxrule=1pt,
  arc=3pt,
  boxsep=5pt,
  left=6pt,  
  enhanced jigsaw  
}

\newtcolorbox{promptbox7}{
  colback=blue!5!white,
  colframe=cyan!30!blue,
  fonttitle=\bfseries,
  title=Prompt for IFEval benchmark,
  boxrule=1pt,
  arc=3pt,
  boxsep=5pt,
  left=6pt,  
  enhanced jigsaw  
}

\newtcolorbox{promptbox8}{
  colback=blue!5!white,
  colframe=cyan!30!blue,
  fonttitle=\bfseries,
  title=Prompt for MedMCQA benchmark,
  boxrule=1pt,
  arc=3pt,
  boxsep=5pt,
  left=6pt,  
  enhanced jigsaw  
}

\newcommand{\ccite}[2][red]{\textcolor{#1}{\cite{#2}}}
\newtcolorbox{promptbox9}{
  colback=blue!5!white,
  colframe=cyan!30!blue,
  fonttitle=\bfseries,
  title=Prompt for Aggregator,
  boxrule=1pt,
  arc=3pt,
  boxsep=5pt,
  left=6pt,  
  enhanced jigsaw  
}

\begin{figure}[th]  
\centering

\begin{promptbox9}
\textbf{System Prompt:} "You have been provided with a set of responses from various open-source models to the latest user query. Your task is to synthesize these responses into a single, high-quality response. It is crucial to critically evaluate the information provided in these responses, recognizing that some of it may be biased or incorrect. Your response should not simply replicate the given answers but should offer a refined, accurate, and comprehensive reply to the instruction. Ensure your response is well-structured, coherent, and adheres to the highest standards of accuracy and reliability.  \\
Responses from models: \\
1.\{Response1\} \\
2.\{Response2\} \\
...
" \\
\textbf{User Prompt:} "Question: \{question\}."
\end{promptbox9}

\vspace{-2mm}
\caption{Prompt for Aggregator within our CCTS-MM}
\label{prompt_aggregator}
\end{figure}



\end{document}